\documentclass{article}
\usepackage{tabularx}
\usepackage{graphicx}
\PassOptionsToPackage{numbers,compress}{natbib}
\usepackage[preprint]{neurips_2026}

\usepackage[utf8]{inputenc} 
\usepackage[T1]{fontenc}    
\usepackage{hyperref}       
\usepackage{url}            
\usepackage{booktabs}       
\usepackage{amsfonts}       
\usepackage{amssymb}        
\usepackage{nicefrac}       
\usepackage{microtype}      
\usepackage{xcolor}         
\usepackage{placeins}       
\usepackage{float}          

\newcommand{\cmark}{\checkmark}
\newcommand{\xmark}{$\times$}
\newif\ifshowoutlinebullets
\showoutlinebulletsfalse 
\newcommand{\outlinebullets}[1]{%
\ifshowoutlinebullets
\FloatBarrier
\begin{samepage}
\paragraph{Outline bullet points}
{\setlength{\itemsep}{0pt}%
\setlength{\parskip}{0pt}%
\setlength{\parsep}{0pt}%
\setlength{\topsep}{2pt}%
\begin{itemize}
#1
\end{itemize}}
\end{samepage}
\FloatBarrier
\fi
}

\title{A Systematic Evaluation of Molecular Mixture Behavior Prediction}

\author{%
  Roel J.~Leenhouts \\
  KU Leuven \\
  Leuven, Belgium \\
  \And
  Nathan K.~Morgan \\
  MIT \\
  Cambridge, USA \\
  \AND
  William H.~Green \\
  MIT \\
  Cambridge, USA \\
  \And
  Jan G.~Rittig\thanks{Corresponding authors:
    \texttt{jan.rittig@epfl.ch} and
    \texttt{florence.vermeire@kuleuven.be}.} \\
  RWTH Aachen University \\
  Aachen, Germany \\
  \And
  Florence H.~Vermeire\footnotemark[\value{footnote}] \\
  KU Leuven \\
  Leuven, Belgium \\
}

\begin{document}

\maketitle

\begin{abstract}
Machine learning for molecular property prediction has focused largely on pure compounds, even though many practical applications depend on mixtures with intermolecular interactions. Recent work has expanded the availability of mixture datasets, but evaluation still focuses mainly on absolute accuracy. However, absolute errors in mixtures conflate pure-component contributions with deviations from ideal mixing.
We propose an evaluation framework that decomposes mixture-property error into pure-compound and interaction (non-ideal) components. The framework combines leakage-aware split protocols, ideal-mixture baselines, and excess-property metrics. To support reproducible benchmarking, we curate seven matched pure and mixture physicochemical property datasets.
Across multiple mixture-property tasks and model families, we find that strong absolute accuracy can mask poor recovery of non-ideal mixture behavior, and that performance drops substantially under strict molecule splits. These results identify transfer to unseen molecules as a central challenge in molecular mixture machine learning and motivate evaluation beyond absolute accuracy alone.
\end{abstract}

\section{Introduction}
\outlinebullets{
\item molecular mixtures as the relevant setting for many practical molecular-property prediction problems;
\item application motivation from solvents, fuels, and related industrial mixture settings;
\item the gap in current mixture-machine-learning evaluations, which focus mainly on absolute accuracy rather than mixture-specific behavior;
\item excess properties from chemical engineering as the key conceptual basis for evaluating non-ideal mixture behavior in machine learning;
\item the role of matched pure and mixture datasets, strict split protocols, an ideal-mixture reference baseline, and metrics for non-ideal mixture behavior in the proposed systematic model evaluation;
\item Figure~\ref{fig:evaluation-overview} as a summary of the datasets, splits, baselines, and metrics used throughout the paper.
}

\noindent
Many molecular property prediction problems involve mixtures in addition to pure compounds. In applications such as reaction and separation processes, performance is governed by intermolecular interactions (e.g., non-ideal mixture behavior) rather than isolated molecular properties \cite{Crystallization, fuel_motivation, PATEL201624}. Predicting these interaction effects is therefore central to practical mixture modeling.
This is especially relevant in industrial settings where performance depends on combining components effectively. For example, in the pharmaceutical industry, solvents contribute 25--100 kg of hazardous waste per kilogram of product \cite{efactor, solvent_waste}, yet only a limited number of solvents are economically and regulatory viable \cite{Clarke_2018}. As a result, improved use of solvent mixtures, rather than discovery of new pure compounds, represents a key lever for reducing waste and improving sustainability \cite{Bao_Tom_Cheng_Aspuru-Guzik_Allen_2024}. A similar case appears for fuel blending, where the maximum biofuel content, such as ethanol in gasoline, must be chosen to satisfy engine-performance requirements \cite{diesel_fuel_blends}.

Mixtures have long been studied in chemical engineering \cite{wilson_and_deal, Fredenslund_Jones_Prausnitz_1975, Klamt_1995}, and recent work has begun to adopt machine learning (ML) methods and conventions for mixture property prediction \cite{SPECHT202419777, VERMEIRE2021129307, Bao_Tom_Cheng_Aspuru-Guzik_Allen_2024, rittig_thermodynamics_2024}. However, the field lacks standardized machine learning evaluation resources and protocols. In contrast, existing molecular property evaluation resources overwhelmingly focus on pure compounds \cite{moleculenet, huang2021therapeutics, qm9, guacamol, ogb,sample_bench}. While several recent datasets have expanded coverage to mixtures \cite{LEENHOUTS2025162232, Miscible_solvents, LEENHOUTS2025133218, BILODEAU2023142454, Kuzhagaliyeva_Horváth_Williams_Nicolle_Sarathy_2022, Malikov_Krasnov_Kiseleva_Meshcheriakova_Kuznetsov_Elistratov_Vasiyarov_Tatarin_Bezzubov_2025}, current evaluations primarily emphasize absolute prediction accuracy \cite{rajaonson2025chemixhub}. For molecular mixtures, however, absolute error mixes two distinct aspects of model performance: how well the pure-component contributions are predicted, and how well the deviation from ideal mixing is captured. As a result, a large share of the error may be determined by pure-compound prediction quality, while mixture-specific behavior is evaluated only indirectly. 

To address this gap, this work frames molecular mixture modeling as an evaluation problem centered on mixture behavior. In standard chemical engineering theory, excess properties quantify deviations from ideal mixing \cite{Prausnitz}. This makes them a natural evaluation lens for molecular mixtures: they distinguish errors arising from pure-component prediction and errors arising from mixture interactions. Conventional models \cite{Renon_Prausnitz_1968, excess_viscosity} and recent machine learning \cite{Liu_2019, rittig_thermodynamics_2024, SPECHT202419777} work have already used excess-property formulations in thermochemical mixture modeling (e.g. vapor-liquid equilibria), but a broader evaluation framework that uses excess properties across physicochemical tasks is still missing. We therefore use excess properties to define physically meaningful ideal-mixture baselines and metrics for non-ideal behavior alongside standard absolute predictive metrics.

Applying this idea in machine learning requires matched pure-compound and mixture datasets. The pure-compound data are needed to construct ideal-mixture references, while the mixture data are needed to measure how well models recover the non-ideal contribution beyond those references. To enable this evaluation, we curate seven publicly available datasets, with properties available in both pure and mixture settings.
The evaluation also requires split protocols that reflect the structure of mixture data. As in molecular machine learning more broadly, split design determines the kind of generalization being tested, and for mixture datasets naive random splits leak the same component combination across train and test through different compositions \cite{Deng_2023, Joeres_2025}.
We therefore define structured split families that separate random interpolation from extrapolation to unseen molecules or their unseen combinations, and complement them with a simple but informative ideal-mixture baseline. Beyond dataset and metric design, the evaluation is used to benchmark mixture ML models, including component featurization, state dependence, and architectural choices for interaction and aggregation.

This evaluation is designed to assess whether state-of-the-art model architectures also capture molecular mixture behavior besides achieving strong absolute accuracy. Our main contributions are:
{\setlength{\itemsep}{0pt}
 \setlength{\parskip}{0pt}
 \setlength{\parsep}{0pt}
 \setlength{\topsep}{0pt}
 \setlength{\partopsep}{0pt}
 \begin{itemize}
  \item standardization and curation of mixture datasets and matched pure-compound datasets;
  \item leakage-aware split protocols for molecular mixture evaluation;
  \item ideal-mixture reference baselines and excess properties for separating pure-compound accuracy from non-ideal mixture behavior;
  \item systematic benchmarking of molecular mixture models across datasets and splits;
\end{itemize}
}
\setlength{\topsep}{0pt}

\section{Related Work}
\paragraph{Molecular evaluations}
\outlinebullets{
\item pure-compound evaluation resources and physically informed comparisons as precedents for molecular evaluation;
\item CheMixHub as the closest prior mixture-focused resource for curated datasets and machine-learning baselines;
\item the remaining gaps in the ideal-mixture reference baseline, metrics for non-ideal mixture behavior, and pure-to-mixture generalization.
}
Evaluation resources for pure compounds, including MoleculeNet \cite{moleculenet}, Therapeutics Data Commons \cite{huang2021therapeutics}, GuacaMol \cite{guacamol}, and QM9 \cite{qm9}, have established the role of standardized datasets and protocols in molecular ML. Related work has highlighted the value of physically informed baselines, and uncertainty-aware evaluation beyond black-box accuracy comparisons \cite{PhysicalPooling, Dobbelaere_2024, Heid_2023}. For molecular mixtures, CheMixHub \cite{rajaonson2025chemixhub} is the closest prior evaluation resource, providing a benchmark with curated datasets and machine-learning comparisons across several mixture-property tasks. However, it does not include an ideal-mixture reference baseline, metrics for non-ideal mixture behavior, or systematic evaluation of pure-to-mixture generalization.

\paragraph{Molecular mixture architectures}
\outlinebullets{
\item interaction-based models for molecular mixtures;
\item aggregation-based models for molecular mixtures;
\item component featurization, and state dependence in mixture representations;
}
Machine learning approaches for molecular mixtures differ primarily in how they represent intermolecular effects. Interaction-based models explicitly exchange information between components, for example through cross-molecular message passing or attention interaction terms \cite{D2DD00045H, attention_accoef, smile_is_all_you_need, rittig2026molecular, gibbs_duhem_gnn}. These models are designed to capture non-ideal behavior more directly. On the other hand, aggregation-based models instead represent a mixture as a set of component embeddings and aggregate them with permutation invariant operations \cite{LEENHOUTS2025133218, LEENHOUTS2025162232, rajaonson2025chemixhub, Kuzhagaliyeva_Horváth_Williams_Nicolle_Sarathy_2022, PRXEnergy.3.023006, DICAPRIO2025110410, cmc_mixtures_gnn, Hanaoka}.

Mixture representations also depend on how individual components are featurized and how state variables such as temperature are handled. Component representations may be based on fixed descriptors \cite{BpKelley, mordred, ecfp}, pretrained molecular features \cite{molt5, molbert}, or learned graph-based embeddings \cite{dmpnn, gilmer, chemprop}. State dependence is handled either through direct machine-learning inputs or through physically motivated equations \cite{jirasek, gh_gnn} that aim to improve robustness and extrapolation.
Despite these modeling differences, these architectural choices have typically been tested on narrower sets of tasks and compared using absolute prediction accuracy alone. Systematic comparison under metrics for non-ideal mixture behavior, an ideal-mixture reference baseline, and stricter generalization settings remains limited.

\section{Methods}
\subsection{Datasets}
\outlinebullets{
\item the overall dataset scope and supported property families;
\item the role of paired pure and mixture datasets in the evaluation;
\item the function of the main dataset table and pure-data coverage table;
}
We study neutral organic small-molecule mixtures across multiple physicochemical properties, varying data availability, and experimental conditions. The datasets include solvation free energy ($\Delta G_{solv}$) \cite{VERMEIRE2021129307, LEENHOUTS2025162232}, vaporization enthalpy ($\Delta H_{vap}$) \cite{Miscible_solvents}, solubility ($\log(S)$) \cite{Malikov_Krasnov_Kiseleva_Meshcheriakova_Kuznetsov_Elistratov_Vasiyarov_Tatarin_Bezzubov_2025, Krasnov_Malikov_Kiseleva_Tatarin_Bezzubov_2025}, viscosity ($\ln(\eta)$) \cite{LEENHOUTS2025133218, BILODEAU2023142454, Larsson_Vermeire_Verhelst_2023}, and fuel performance indicators (derived cetane number, DCN, and motor octane number, MON) \cite{LEENHOUTS2025133218, Kuzhagaliyeva_Horváth_Williams_Nicolle_Sarathy_2022, Larsson_Vermeire_Verhelst_2023}. We describe the curation and standardization pipeline and provide property definitions in the appendix. The curated datasets and fixed split definitions are publicly available on Zenodo (\url{https://doi.org/10.5281/zenodo.19914760})
Table~\ref{tab:dataset-summary-mix} summarizes the mixture datasets, while Appendix Table~\ref{tab:dataset-summary-pure} reports the corresponding pure-compound datasets used for ideal-mixture references and excess-property calculations. Additional distribution summaries are provided in the Appendix Figures~\ref{fig:mix-value-mw-dist}, \ref{fig:component-percentage}, and \ref{fig:excess-value-violin-distributions}.

Most data points correspond to binary to quinary mixtures, with MON as the main outlier exhibiting substantial coverage of higher-order mixtures with up to 121 components. Pure-reference coverage is high for most datasets, enabling excess-property metrics to be computed for a large fraction of the benchmark.
Compared with standard pure-compound benchmarks \cite{huang2021therapeutics}, the number of unique molecules is relatively small; benchmark difficulty is therefore driven primarily by combinatorial mixture structure rather than broad molecular diversity.

\begin{table}[h]
\centering
\small
\caption{Overview of the standardized mixture datasets.  \# Unique mix counts distinct component combinations after canonicalizing component order and ignoring fractions and row context. \# Unique mol counts distinct mixture-component molecule identities and excludes solutes where applicable. \% Pure coverage reports the fraction of mixture rows for which the required pure-compound reference is available, including temperature and solute context when present.}
\label{tab:dataset-summary-mix}
\setlength{\tabcolsep}{4pt}
\resizebox{\textwidth}{!}{%
\begin{tabular}{lccrrrccl}
\hline
Task & Type & Data points & \# Unique mix & \# Unique mol & Context & \% Pure coverage & Unit & Source \\
\hline
$\Delta G_{solv}$ & Comp & 422669 & 452 & 32 & solute & 38.6\% & kcal/mol & \cite{LEENHOUTS2025162232} \\
$\log(S)$ & Exp & 125932 & 726 & 132 & solute, T & 75.3\% & - & \cite{ Malikov_Krasnov_Kiseleva_Meshcheriakova_Kuznetsov_Elistratov_Vasiyarov_Tatarin_Bezzubov_2025} \\
$\Delta H_{vap}$ & Comp & 30061 & 19157 & 81 & - & 100.0\% & kcal/mol & \cite{Miscible_solvents} \\
$\ln(\eta)$ & Exp & 35938 & 3377 & 564 & T & 88.1\% & ln(mPa$\cdot$s) & \cite{BILODEAU2023142454, LEENHOUTS2025133218} \\
$T_{flash}$ & Exp & 1006 & 113 & 71 & - & 96.7\% & $^{\circ}$C & \cite{LEENHOUTS2025133218} \\
$DCN$ & Exp & 484 & 57 & 35 & - & 74.4\% & - & \cite{LEENHOUTS2025133218} \\
$MON$ & Exp & 292 & 67 & 131 & - & 73.6\% & - & \cite{ Kuzhagaliyeva_Horváth_Williams_Nicolle_Sarathy_2022} \\
\hline
\end{tabular}%
}
\end{table}

\subsection{Evaluation Protocols}

\subsubsection{Data splits}
\outlinebullets{
\item the structured split families used throughout the evaluation;
\item the limitations of random row-wise splits for repeated mixture combinations;
\item pure-to-mixture, unseen combination, unseen molecule, and temperature-tail extrapolation settings;
}

For evaluation, four structured split families that target distinct generalization settings are applied. Table~\ref{tab:split-protocol-overview} summarizes which information remains available during training for each split family. In all splits, available pure compound data is included in the train partition, and explicitly removed from the test partition.
As a comparison point, random splits are considered, but the main evaluation focuses on the structured splits.

\begin{table}[h]
\centering
\small
\caption{Evaluation splits. Check marks show which information remains available during training.}
\label{tab:split-protocol-overview}
\setlength{\tabcolsep}{8pt}
\renewcommand{\arraystretch}{0.85}
\begin{tabular}{@{}p{2.7cm}>{\centering\arraybackslash}p{2.2cm}>{\centering\arraybackslash}p{2.2cm}>{\centering\arraybackslash}p{2.6cm}>{\centering\arraybackslash}p{2.0cm}@{}}
\toprule
Split & Molecule types in train & Mixture info in train & Molecule combination in train & Temperature info in train \\
\midrule
Random & \cmark & \cmark & \cmark & \cmark \\
Mixture & \cmark & \cmark & \xmark & \cmark \\
Mixture-temperature & \cmark & \cmark & \xmark & \xmark \\
Molecule & \xmark & \cmark & \xmark & \cmark \\
Pure-to-mixture & \cmark & \xmark & \xmark & \cmark \\
\bottomrule
\end{tabular}
\end{table}

\paragraph{Random splits.}
Rows are assigned to train, validation, and test sets independently at random. The same molecule identities and mixture combinations can therefore still appear across partitions at different compositions or conditions, herein temperature.
\paragraph{Mixture splits.}
Specific component combinations are held out, while individual molecule identities may still appear elsewhere in training. This tests generalization to unseen combinations rather than unseen components.
\paragraph{Mixture-temperature splits.}
Mixture-temperature splits are derived from the mixture split by imposing temperature extrapolation. For each original mixture split fold, two settings are constructed: training and validation are restricted to one half of the temperature range (lower or upper 50\%), while testing is restricted to the opposite extreme (upper or lower 10\% tail, respectively).
\paragraph{Molecule splits.}
Test mixtures contain solely held-out molecule identities, so the model must generalize to mixtures with completely unseen components. Rows that mix held-out and retained molecules are excluded from both train and test sets to avoid leakage across the train--test boundary. For DCN and MON, the combination of small dataset size and broad molecule overlap leaves too few feasible held-out mixtures, so this split is not used there.
\paragraph{Pure-to-mixture split.}
This split uses the same test set as the molecule split but trains only on pure-compound data, constituting a transfer learning setting that isolates how well models transfer single-molecule knowledge to mixture behavior.

\subsubsection{Evaluation Metrics and Baselines}
\begin{figure}[h]
    \centering
    \includegraphics[width=0.43\linewidth]{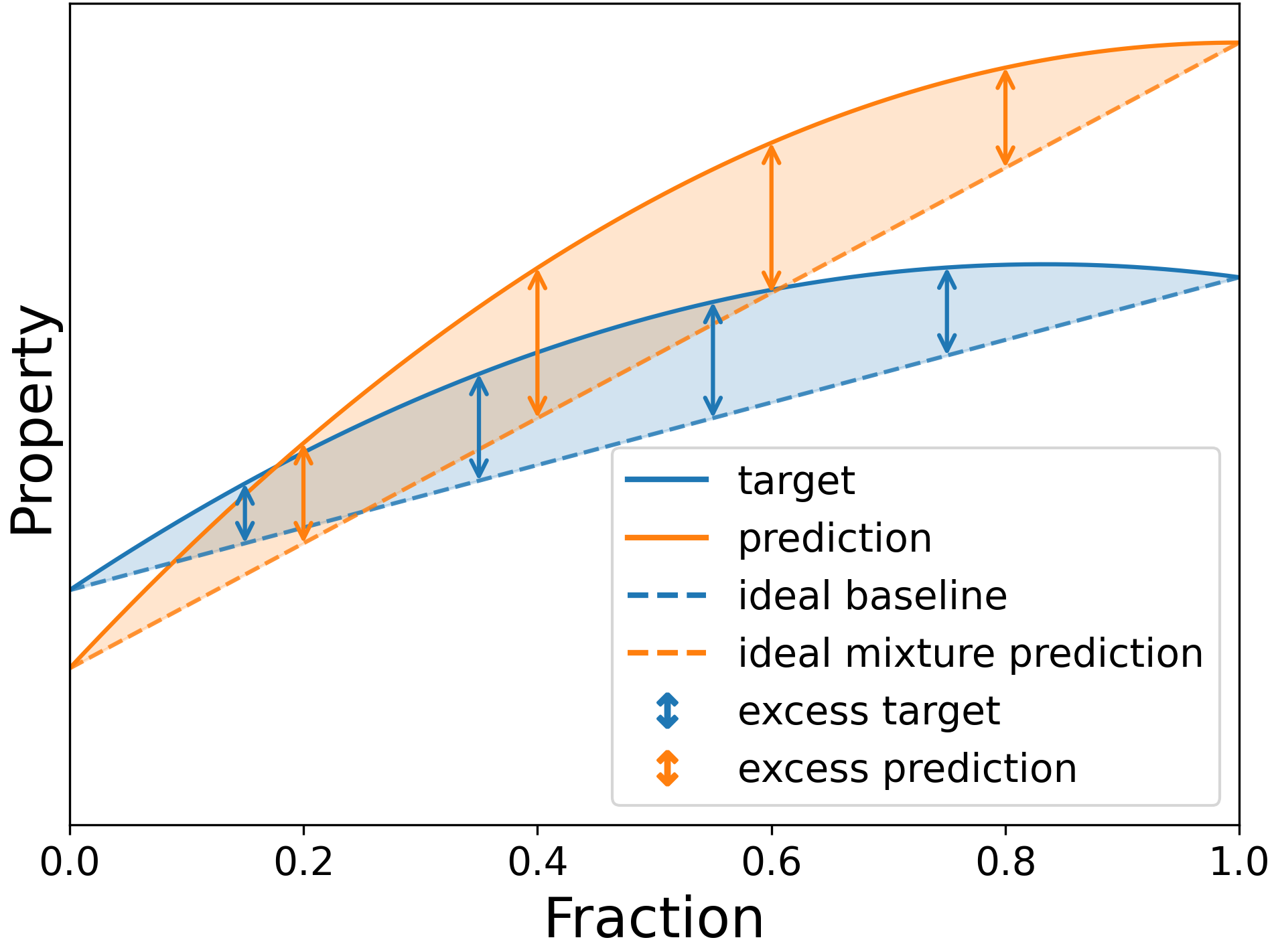}
    \hfill
    \includegraphics[width=0.43\linewidth]{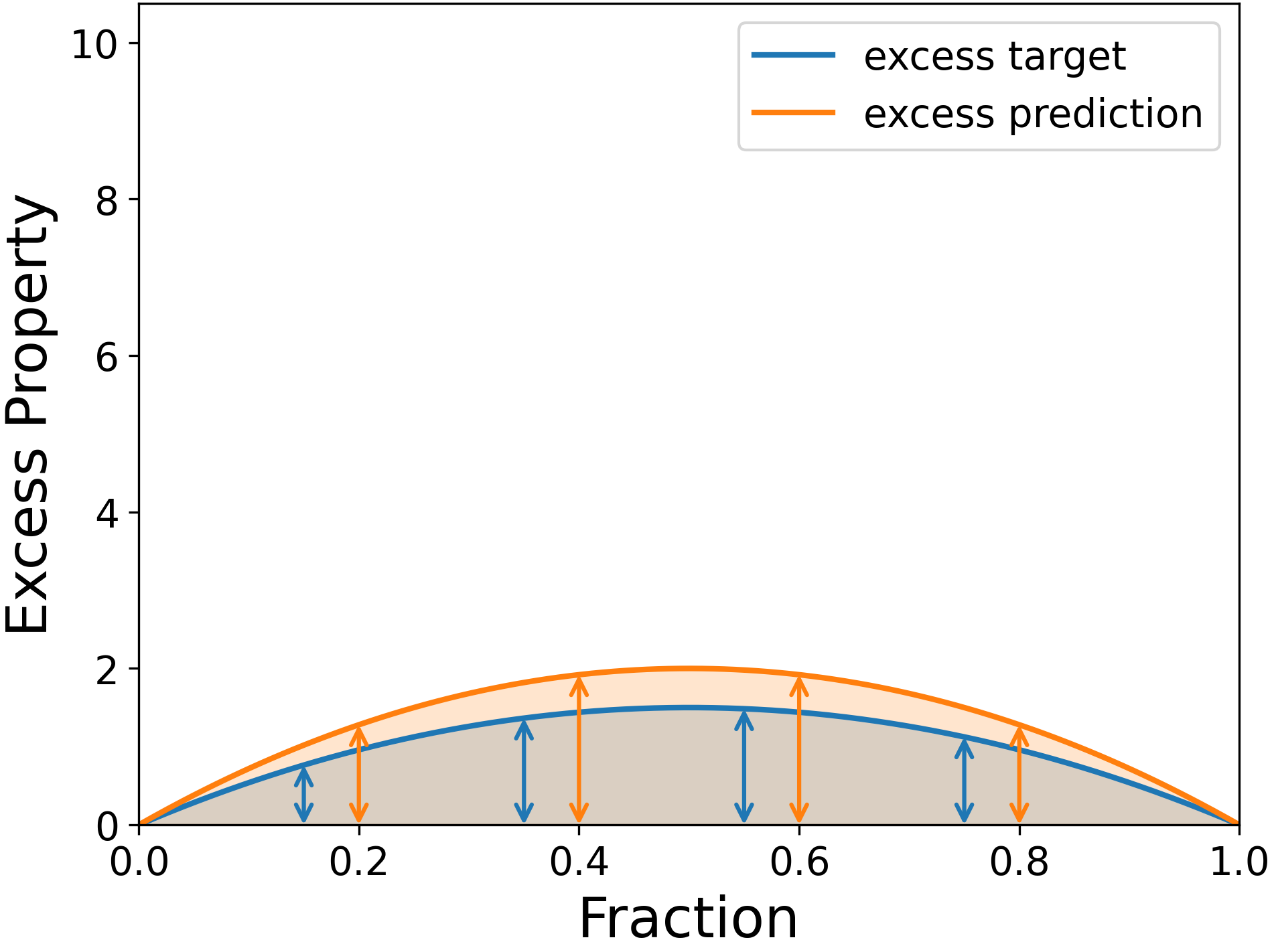}
    \caption{Left: target and predicted mixture-property curves, with shaded regions indicating excess properties. Right: the corresponding excess-property, isolating deviations from the ideal reference.}
    \label{fig:fill-patterns}
\end{figure}

\outlinebullets{
\item excess properties and trend-based metrics for mixture behavior;
\item the thermodynamic definition of ideal and excess mixture properties;
\item the ideal-mixture reference baseline;
}
\paragraph{Excess properties and metrics for non-ideal mixture behavior.}
We complement absolute error metrics with metrics that more directly assess mixture behavior.
First, we compute \emph{excess properties}, defined in thermodynamics as the deviation of a real-mixture property from its ideal-mixture value under the same conditions (e.g. temperature and solute) \cite{Prausnitz}. Figure~\ref{fig:fill-patterns} illustrates the ideal and excess property decomposition for an example mixture combination. For a mixture property \(z\),
\begin{equation}
    \begin{array}{rcl}
        z^{E} & = & z - z^{\mathrm{id}}, \qquad
        z^{\mathrm{id}} = \sum_{i=1}^{N} x_i z_i^{\mathrm{pure}}.
    \end{array}
\end{equation}
Here \(z^{\mathrm{id}}\) denotes the corresponding ideal-mixture property, \(x_i\) is the fraction of component \(i\) in an \(N\)-component mixture, and \(z_i^{\mathrm{pure}}\) is the pure-compound property of component \(i\) at the matched condition. The ground-truth and predicted excess properties are therefore
\begin{equation}
    z_{\mathrm{true}}^{E} = z_{\mathrm{true}} - z^{\mathrm{id}},
    \qquad
    \hat{z}^{E} = \hat{z} - \hat{z}^{\mathrm{id}},
\end{equation}

The appropriate ideal reference is property-dependent: the linear form is physically motivated for \(\Delta G_{solv}\) and \(\Delta H_{vap}\), applied on the transformed scales \(\log(S)\) and \(\ln(\eta)\) for solubility and viscosity, and used only as a heuristic composition-weighted reference for \(T_{flash}\), \(DCN\), and \(MON\).
Predicted excess properties are computed relative to the model-implied ideal mixture value, \(\hat{z}^{\mathrm{id}}\), so that the predicted excess reflects the model's own decomposition into ideal and non-ideal contributions.

Second, to assess whether models recover correct compositional trends, we report the average Kendall rank correlation computed within each mixture combination. This trend-based metric measures whether models recover the correct compositional ordering of mixture properties.

\paragraph{Ideal mixture reference baseline.}
We introduce a baseline as reference for model comparison based on the ideal-mixture approximation, computing mixture properties via composition-weighted interpolation of matched pure-compound values. This is thermodynamically grounded for $\Delta G_{solv}$, $\Delta H_{vap}$, $\log(S)$, and $\ln(\eta)$, and used as a heuristic baseline for flash point, DCN, and MON.

\subsection{Models}
\begin{figure}[h]
    \centering
    \includegraphics[width=1.0\linewidth]{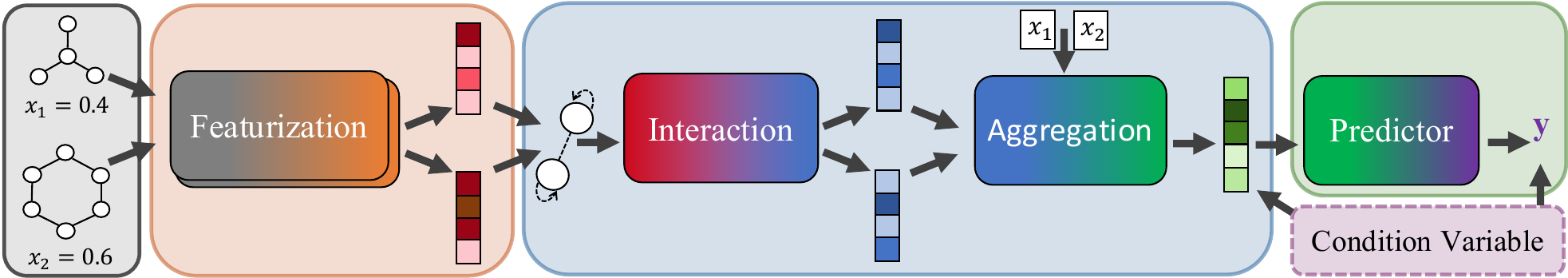}
    \caption{Overview of the four model comparison axes: component featurization (orange), interaction modules (blue), predictors (green), and condition variables (purple).}
    \label{fig:evaluation-overview}
\end{figure}

Model configurations are compared along four axes: interaction and pooling design, component featurization, prediction heads, and thermodynamic condition variables. All variants were implemented within the same adapted Chemprop framework \cite{chemprop, chempropv2}, allowing architectural choices to be compared under a shared implementation. 
Code and reproducibility materials for this evaluation are available at \url{https://gitlab.kuleuven.be/creas/vermeiregroup/mixture_evaluation}. Figure~\ref{fig:evaluation-overview} summarizes the model design space.

\paragraph{Component featurization.}
We compare three classes of component featurization: task-specific learned graph embeddings, pretrained molecular features, and fixed chemoinformatics descriptors. Learned graph-based embeddings are obtained with directed message passing neural networks (D-MPNN) \cite{gilmer, dmpnn}. As a pretrained representation, we include MolT5-based molecular features \cite{molt5} as implemented through MolFeat \cite{molfeat}. As a fixed feature representation, we include 200-dimensional normalized RDKit descriptors \cite{Landrum} computed using Descriptastorus \cite{BpKelley}.

\paragraph{Interaction modules and aggregation functions.}
Within the shared graph-based implementation, we vary both the interaction module and the aggregation function used to construct mixture representations. For interaction modeling, we compare molecular message passing \cite{D2DD00045H}, interaction message passing \cite{gh_gnn}, and no explicit interaction block. For aggregation, we compare weighted-sum, concatenation, DeepSets \cite{Zaheer_2017}, attentive \cite{dosSantos_2016}, and Set2Set \cite{Vinyals_2016}. For MolT5 and descriptor features, we evaluate the compatible subset of these mixture-construction choices, reusing aggregation without graph interaction blocks.

\paragraph{Predictors.}
Standardized predictors are used for the final regression stage. For neural models, the aggregated mixture embedding is passed to a multilayer perceptron (MLP) to produce property predictions. 
For descriptor-based models, we additionally use gradient-boosted decision trees (XGBoost) \cite{xgboost} on composition-aggregated descriptor features.

\paragraph{Thermodynamic condition variables.}
For temperature-dependent tasks, temperature is treated as an explicit state variable. We use omission of temperature as the baseline setting, and compare it against two inclusion strategies: concatenating a temperature feature to the learned mixture representation, and a physics-motivated temperature variant. For the latter, temperature enters through the Van 't Hoff equation for solubility and the Arrhenius equation for viscosity \cite{rajaonson2025chemixhub}.

\paragraph{Training procedure.}
Details on the training workflow, including hyperparameter optimization setup, are provided in~\ref{app:training-hparams}.
In general, the evaluation proceeds in two stages. We first compare the three model families under a common reference configuration, using weighted-sum aggregation together with a fixed set of hyperparameters selected from cross-dataset runs. The goal of this stage is to identify a single parameter setting that remains competitive across the physicochemical property datasets within the scope of this benchmark, rather than re-tuning separately for each task. We then study architectural variations around this reference setup, including interaction modules, aggregation choices, and condition variable strategies. Unless stated otherwise, all reported metrics are computed as the mean and sample standard deviation over the five cross-validation folds.

\section{Results \& Discussion}

\subsection{Evaluating pure-to-mixture generalization}
\outlinebullets{
\item model performance under absolute, excess-property, and trend-based evaluation;
\item cases where absolute accuracy masks weak recovery of non-ideal mixture behavior.
}
\begin{figure*}[h]
    \centering
    \includegraphics[width=1.\textwidth]{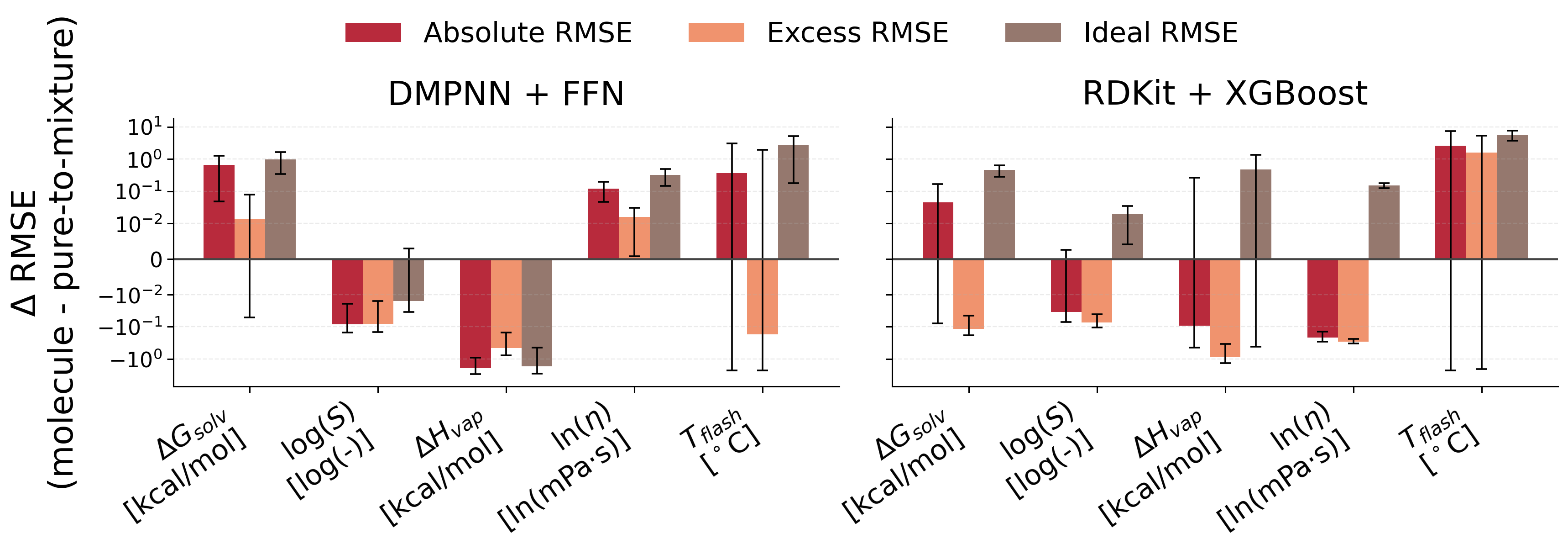}
    \caption{Differences in absolute, excess, and ideal RMSE from pure-to-mixture to molecule splits for the DMPNN + FFN and RDKit + XGBoost model families.}
    \label{fig:matrix1-pure-to-mixture-vs-molecule-type-rmse}
\end{figure*}

Figure~\ref{fig:matrix1-pure-to-mixture-vs-molecule-type-rmse} compares the $\Delta$RMSE between the pure-to-mixture and molecule splits for the DMPNN + FFN and RDKit + XGBoost model families; corresponding MolT5 + FFN results are provided in Appendix Figure~\ref{fig:matrix1-molt5-pure-to-mixture-vs-molecule-type-rmse}.
Pure-to-mixture models are expected to recover ideal contributions more accurately but lack direct exposure to interaction effects, whereas molecule-split models are exposed to mixture behavior but receive less direct supervision on pure-component properties. Across datasets and model families, pure-to-mixture training more frequently yields lower ideal-component error, whereas molecule splits more frequently yield lower excess-property error, though neither trend is universal. Absolute RMSE, however, conflates these two effects. These results demonstrate that absolute metrics can obscure how well a model captures non-ideal mixture behavior, depending on the supervision available during training, and motivate the use of excess metrics for evaluating mixture-behavior recovery.

\begin{table*}[h]
\centering
\scriptsize
\caption{Excess RMSE across data splits. Values are mean $\pm$ std over completed folds. Bold marks the best mean; underlining marks values within two standard deviations of the best mean.}
\label{tab:default-hparam-matrix1-excess-rmse-picture-style}
\setlength{\tabcolsep}{4pt}
\resizebox{\textwidth}{!}{%
\begin{tabular}{llccccc}
\toprule
Split & Model type & \shortstack{$\Delta G_{solv}$\\ {[kcal/mol]}} & \shortstack{$\log(S)$\\ {[log(-)]}} & \shortstack{$\Delta H_{vap}$\\ {[kcal/mol]}} & \shortstack{$\ln(\eta)$\\ {[ln(mPa·s)]}} & \shortstack{$T_{flash}$\\ {[$^\circ$C]}} \\
\midrule
Random & Ideal baseline & 0.71 $\pm$ 0.0064 & 0.77 $\pm$ 0.0073 & 1.0 $\pm$ 0.0082 & 0.30 $\pm$ 0.015 & 14. $\pm$ 0.78 \\
 & DMPNN + FFN & \underline{0.14 $\pm$ 0.0080} & \textbf{0.075 $\pm$ 0.0040} & \textbf{0.066 $\pm$ 0.0048} & \textbf{0.096 $\pm$ 0.0027} & 10. $\pm$ 0.82 \\
 & MolT5 + FFN & \textbf{0.13 $\pm$ 0.0057} & 0.15 $\pm$ 0.049 & 0.14 $\pm$ 0.041 & 0.11 $\pm$ 0.0045 & 11. $\pm$ 0.67 \\
 & RDKit + XGBoost & 0.34 $\pm$ 0.0015 & 0.29 $\pm$ 0.0030 & 0.20 $\pm$ 0.0044 & 0.14 $\pm$ 0.0091 & \textbf{4.3 $\pm$ 0.44} \\
\midrule
Mixture & Ideal baseline & 0.71 $\pm$ 0.053 & 0.74 $\pm$ 0.22 & 1.1 $\pm$ 0.23 & 0.30 $\pm$ 0.035 & \underline{13. $\pm$ 5.4} \\
 & DMPNN + FFN & \textbf{0.18 $\pm$ 0.037} & \textbf{0.36 $\pm$ 0.062} & \textbf{0.11 $\pm$ 0.064} & \textbf{0.18 $\pm$ 0.034} & \underline{11. $\pm$ 4.7} \\
 & MolT5 + FFN & \underline{0.19 $\pm$ 0.034} & \underline{0.37 $\pm$ 0.042} & \underline{0.13 $\pm$ 0.070} & \underline{0.20 $\pm$ 0.042} & \underline{11. $\pm$ 4.7} \\
 & RDKit + XGBoost & 0.39 $\pm$ 0.026 & \underline{0.37 $\pm$ 0.057} & 0.29 $\pm$ 0.16 & \underline{0.21 $\pm$ 0.048} & \textbf{9.1 $\pm$ 3.2} \\
\midrule
Molecule & Ideal baseline & \underline{0.65 $\pm$ 0.23} & \underline{0.81 $\pm$ 0.39} & 0.87 $\pm$ 0.33 & \underline{0.21 $\pm$ 0.062} & \underline{13. $\pm$ 8.3} \\
 & DMPNN + FFN & \textbf{0.52 $\pm$ 0.17} & \underline{0.74 $\pm$ 0.33} & \textbf{0.50 $\pm$ 0.15} & \underline{0.24 $\pm$ 0.064} & \textbf{13. $\pm$ 6.1} \\
 & MolT5 + FFN & \underline{0.59 $\pm$ 0.20} & \underline{0.80 $\pm$ 0.37} & \underline{0.57 $\pm$ 0.18} & \textbf{0.21 $\pm$ 0.061} & \underline{15. $\pm$ 7.1} \\
 & RDKit + XGBoost & \underline{0.60 $\pm$ 0.14} & \textbf{0.68 $\pm$ 0.32} & 0.80 $\pm$ 0.20 & \underline{0.26 $\pm$ 0.050} & \underline{17. $\pm$ 6.1} \\
\midrule
Pure-to-mixture & DMPNN + FFN & \textbf{0.51 $\pm$ 0.17} & \underline{0.83 $\pm$ 0.39} & \underline{0.96 $\pm$ 0.29} & \textbf{0.23 $\pm$ 0.057} & \underline{13. $\pm$ 7.4} \\
 & MolT5 + FFN & \underline{0.61 $\pm$ 0.20} & \underline{0.85 $\pm$ 0.39} & \textbf{0.63 $\pm$ 0.30} & \underline{0.24 $\pm$ 0.067} & \textbf{13. $\pm$ 8.9} \\
 & RDKit + XGBoost & \underline{0.71 $\pm$ 0.18} & \textbf{0.76 $\pm$ 0.31} & 1.7 $\pm$ 0.48 & 0.55 $\pm$ 0.076 & \underline{16. $\pm$ 4.0} \\
\bottomrule
\end{tabular}%
}
\end{table*}

Table~\ref{tab:default-hparam-matrix1-excess-rmse-picture-style} reports excess RMSE across the mixture, molecule, and pure-to-mixture splits; Appendix Tables~\ref{tab:default-hparam-matrix1-absolute-rmse-picture-style}, \ref{tab:default-hparam-matrix1-ideal-rmse-picture-style}, and~\ref{tab:default-hparam-matrix1-absolute-kendall-picture-style} provide the corresponding absolute RMSE, ideal RMSE, and Kendall ranking results.
Across splits, performance degrades markedly when moving from the random split to the stricter molecule split, where models must generalize to mixtures composed of entirely unseen molecules. Errors increase across all metrics and models often fail to substantially outperform the ideal-mixture baseline in this setting. While DMPNN + FFN performs strongly on the random and mixture splits, the clearer differences observed on less restrictive splits largely disappear under the molecule split, with no consistently dominant model family. This trend is consistent with the dataset structure: mixture datasets contain fewer unique molecule identities than their paired pure-compound datasets (Appendix Table~\ref{tab:dataset-summary-pure}) and pure-compound benchmarks \cite{moleculenet}, making extrapolation to unseen molecules substantially more challenging and highlighting the difficulty of molecular extrapolation in current mixture-learning benchmarks.

Comparing Tables~\ref{tab:default-hparam-matrix1-excess-rmse-picture-style} and~\ref{tab:default-hparam-matrix1-absolute-rmse-picture-style} for the molecule split highlights the distinction between absolute and excess error relative to the ideal-mixture baseline. For excess RMSE, the ideal baseline corresponds to predicting zero excess (i.e. a horizontal line). While this baseline can outperform models under absolute RMSE, the models achieve substantially lower excess RMSE, indicating that they capture non-ideal mixture behavior beyond trivial predictions. This discrepancy underscores the importance of evaluating excess RMSE: absolute metrics alone can favor baselines that ignore interaction effects.

\subsection{Featurization and predictor comparison}
\begin{figure*}[h]
    \centering
    \includegraphics[width=1.\textwidth]{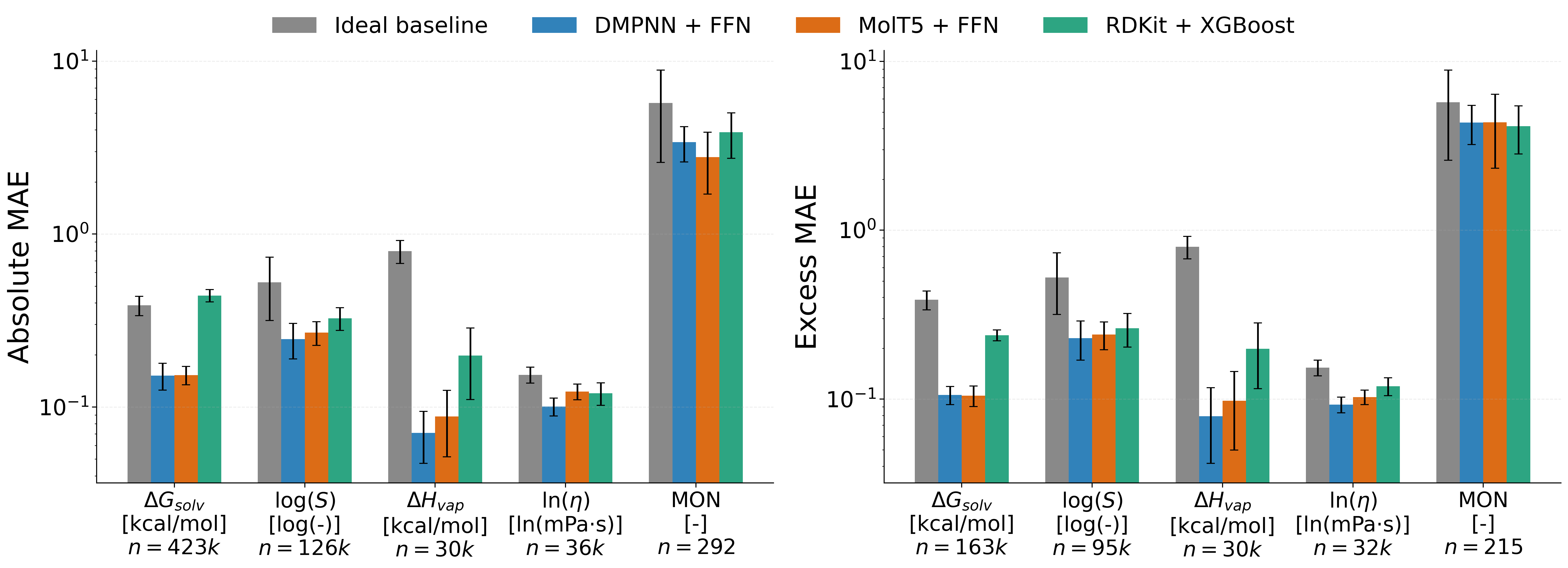}
    \caption{Absolute and excess MAE vs. the ideal-mixture reference baseline under the mixture split.}
    \label{fig:matrix1-mae-vs-ideal-baseline}
\end{figure*}

Figure~\ref{fig:matrix1-mae-vs-ideal-baseline} compares featurization and predictor combinations to the ideal-mixture baseline under the mixture split using absolute and excess MAE (full results in Appendix Figures~\ref{fig:matrix1-absolute-mae-vs-ideal-baseline-all} and~\ref{fig:matrix1-excess-mae-vs-ideal-baseline-all}). In the excess panel, the baseline corresponds to predicting zero excess. Differences between model families are most clearly resolved for the large computational datasets ($\Delta G_{solv}$, $\Delta H_{vap}$), where DMPNN + FFN and MolT5 + FFN attain the lowest mean errors. On large experimental datasets (10K+), differences between model families are small, but all models outperform the ideal-mixture baseline. On smaller experimental datasets, differences are small both among models and relative to the baseline, making model ranking unstable. Overall, featurization and predictor choice is most clearly resolved in high-data computational settings, with DMPNN + FFN and MolT5 + FFN the most consistent best models across datasets.

\begin{figure*}[h]
    \centering
    \includegraphics[width=0.48\linewidth]{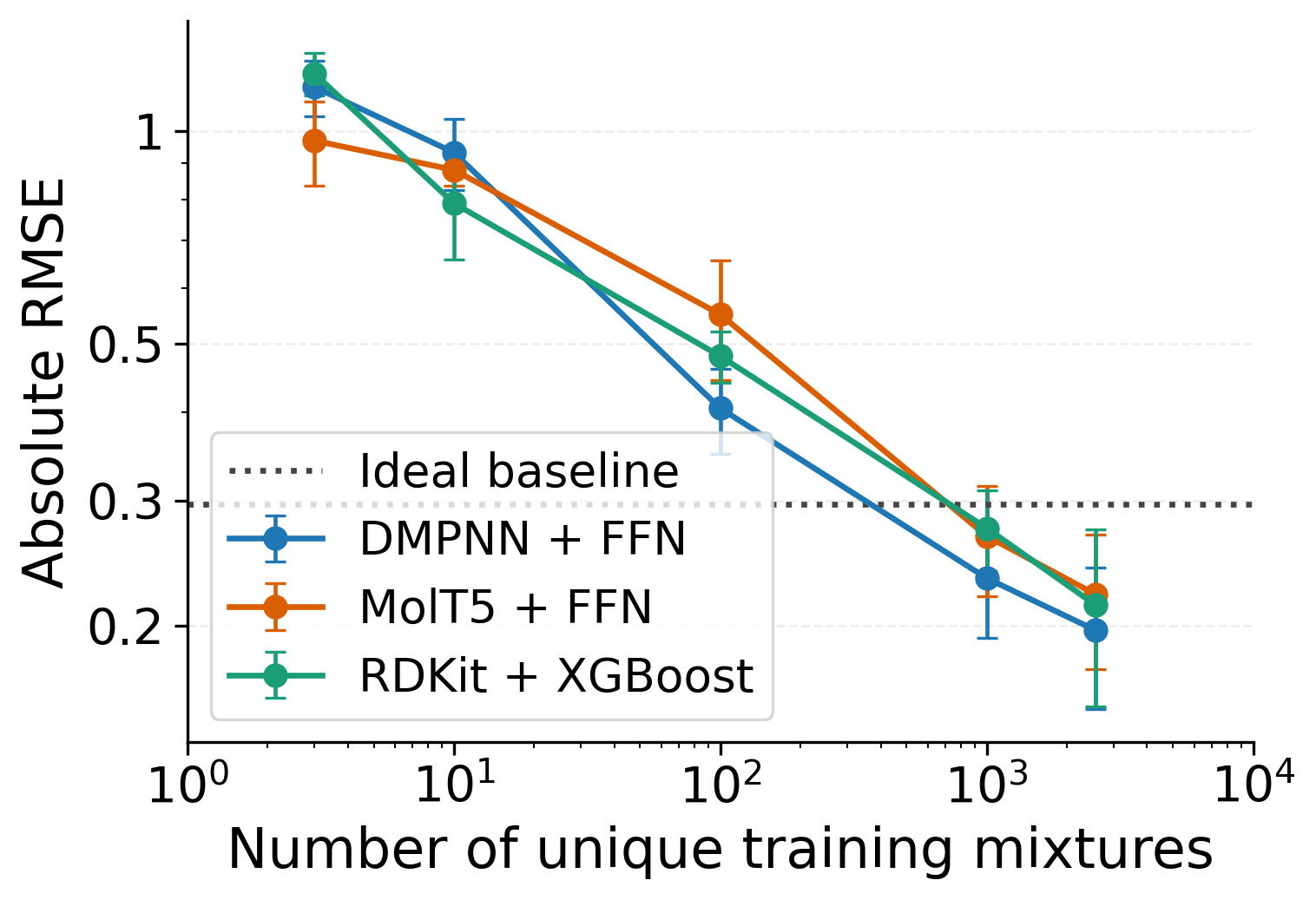}
    \includegraphics[width=0.48\linewidth]{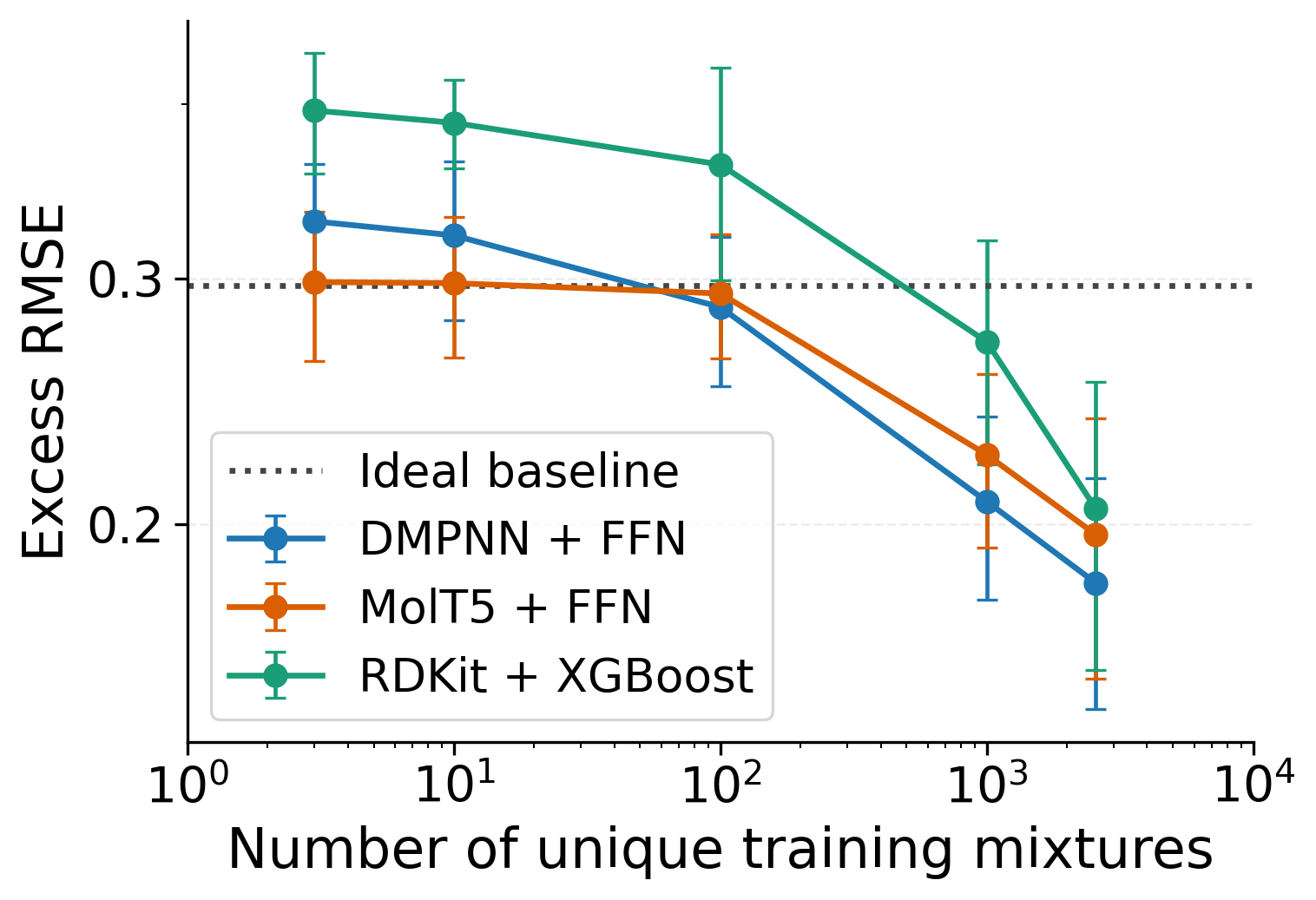}
    \caption{Absolute and excess RMSE learning curves under the mixture split for $\ln(\eta)$, using fixed validation and test sets with a progressively downsampled number of unique training mixtures.}
    \label{fig:matrix5-featurization-learning-curves-viscosity}
\end{figure*}

Figure~\ref{fig:matrix5-featurization-learning-curves-viscosity} shows the viscosity learning curves under the mixture split; the remaining absolute- and excess-RMSE learning curves are provided in Appendix Figures~\ref{fig:matrix5-learning-curves-absolute-appendix} and~\ref{fig:matrix5-learning-curves-excess-appendix}. For viscosity, all model families improve consistently with increasing numbers of unique training mixtures. Across tasks, the excess-RMSE learning curves show that RDKit + XGBoost underperforms across training-set sizes, consistent with the trends observed on the full datasets. In contrast, no consistent performance ordering emerges between MolT5 + FFN and DMPNN + FFN across tasks or data regimes.
The results further suggest that models generally require on the order of $10^3$ unique training mixtures before outperforming the ideal-mixture baseline across properties.
Overall, the learning curves indicate that improvements beyond the ideal-mixture baseline depend strongly on dataset scale, with clear gains in larger datasets and substantially less separable performance in smaller-data regimes.

\subsection{Interaction-module and aggregation comparisons}
\label{sec:interaction-aggregation-results}
\outlinebullets{
\item comparison of mixture-construction choices across interaction modules and aggregation functions;
\item the effect of these architectural choices on generalization and mixture-behavior recovery.
}
Table~\ref{tab:pooling-rmse-2split-compact} reports excess RMSE for DMPNN + FFN models across aggregation schemes, while the corresponding MolT5 + FFN results are provided in Appendix Table~\ref{tab:pooling-rmse-2split-compact-molt5}. Aggregators are evaluated on both the mixture and pure-to-mixture splits. The weighted-sum aggregator imposes a fixed additive structure without learnable parameters, whereas DeepSets, Attentive, and Set2Set use learnable aggregation mechanisms. Mean performance is broadly similar across aggregation modules on both splits. However, each learnable aggregator exhibits at least one task where performance falls more than two standard deviations below the best mean on a split, whereas weighted-sum remains the most consistent and achieves the best mean on six of the seven tasks. Overall, weighted-sum provides the most reliable generalization across tasks and splits.

Interaction modules, \textit{i.e.} MPNN-based components that pass messages between molecular graphs to model cross-component interactions \cite{D2DD00045H, gh_gnn}, are intended to capture non-ideal mixture behavior more directly. However, across datasets, their inclusion does not substantially affect performance. On the mixture split, interaction modules yield no consistent improvement in excess RMSE, suggesting that the additional interaction capacity is either not effectively utilized or unnecessary given the available data.

\begin{table*}[h]
\centering
\scriptsize
\caption{DMPNN + FFN aggregation comparison. Values are excess RMSE $\pm$ std. Bold marks the best per column; underlining marks values within two standard deviations of the best mean.}
\label{tab:pooling-rmse-2split-compact}
\setlength{\tabcolsep}{5pt}
\resizebox{\textwidth}{!}{%
\begin{tabular}{lllcccc}
\toprule
Split & Interaction & Aggregation & $\Delta G_{solv}$ & $\ln(\eta)$ & $T_{flash}$ & MON \\
\midrule
Mixture & None & attentive & \underline{0.22 $\pm$ 0.022} & \underline{0.18 $\pm$ 0.035} & \underline{9.9 $\pm$ 4.8} & \underline{6.4 $\pm$ 2.4} \\
 &  & weighted-sum & \textbf{0.19 $\pm$ 0.048} & \textbf{0.18 $\pm$ 0.035} & \underline{11. $\pm$ 4.7} & \textbf{5.3 $\pm$ 1.1} \\
 &  & DeepSets & 0.71 $\pm$ 0.053 & \underline{0.19 $\pm$ 0.035} & \underline{10. $\pm$ 4.8} & \underline{7.0 $\pm$ 1.1} \\
 &  & Set2Set & \underline{0.21 $\pm$ 0.044} & \underline{0.19 $\pm$ 0.040} & \textbf{9.8 $\pm$ 3.9} & \underline{6.7 $\pm$ 3.0} \\
\midrule
Pure-to-mixture & None & attentive & \underline{0.69 $\pm$ 0.17} & 0.39 $\pm$ 0.032 & \underline{23. $\pm$ 12.} & -- \\
 &  & weighted-sum & \textbf{0.53 $\pm$ 0.17} & \textbf{0.23 $\pm$ 0.056} & \textbf{13. $\pm$ 7.4} & -- \\
 &  & DeepSets & \underline{0.65 $\pm$ 0.23} & \underline{0.25 $\pm$ 0.054} & \underline{14. $\pm$ 7.8} & -- \\
 &  & Set2Set & \underline{0.85 $\pm$ 0.19} & 0.37 $\pm$ 0.016 & \underline{24. $\pm$ 12.} & -- \\
\midrule
Mixture & None & weighted-sum & \textbf{0.19 $\pm$ 0.048} & \textbf{0.18 $\pm$ 0.035} & \underline{11. $\pm$ 4.7} & \textbf{5.3 $\pm$ 1.1} \\
 & Molecular & weighted-sum & 0.34 $\pm$ 0.065 & \underline{0.19 $\pm$ 0.037} & \underline{8.7 $\pm$ 1.9} & 7.8 $\pm$ 3.7 \\
 & Explicit & weighted-sum & \underline{0.27 $\pm$ 0.020} & \underline{0.19 $\pm$ 0.042} & \textbf{7.6 $\pm$ 2.0} & 7.8 $\pm$ 3.7 \\
\bottomrule
\end{tabular}%
}
\end{table*}

\subsection{Temperature modeling}
\begin{figure}[h]
    \centering
    \includegraphics[width=0.95\linewidth]{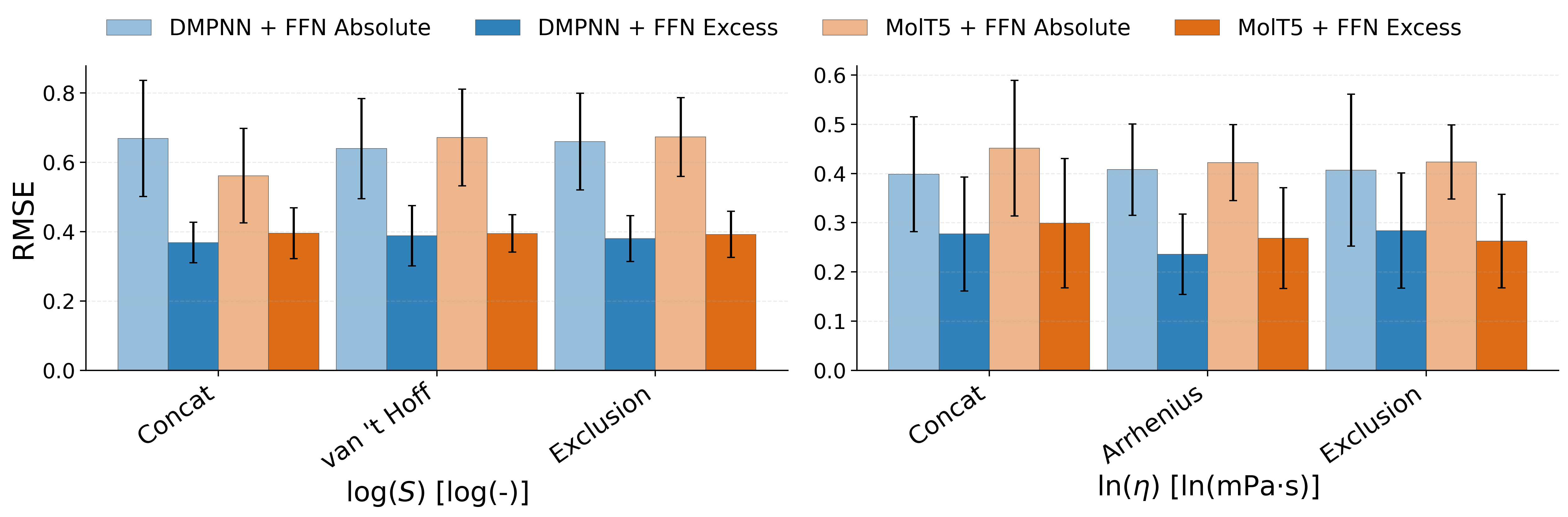}
    \caption{Absolute and excess RMSE for temperature-context variants.}
    \label{fig:temperature-combined-rmse}
\end{figure}

\outlinebullets{
\item model behavior on temperature-dependent tasks and state-aware evaluation settings;
\item the effect of temperature modeling choices on generalization across temperature ranges.
}
Figure~\ref{fig:temperature-combined-rmse} summarizes the temperature-context ablation across two temperature-dependent tasks, reporting absolute RMSE and excess RMSE; Appendix Figure~\ref{fig:temperature-context-fold0-distributions} shows the corresponding train/test temperature distributions. Across temperature-based splits, we observe no consistent advantage for any temperature-handling strategy relative to the no-temperature baseline (Exclusion) in Figure ~\ref{fig:temperature-combined-rmse}. In particular, neither physics-informed heads nor simple feature concatenation consistently outperform omitting temperature.
These results contrast prior work \cite{rajaonson2025chemixhub}, which reported gains from physics-based heads on temperature-bin splits for viscosity. As the same dataset is used, a difference in split design is one plausible contributing factor, though we have not controlled for this directly and other methodological differences may also play a role.
Overall, these findings indicate that the benefit of temperature-aware modeling is sensitive to evaluation design and does not consistently translate under stricter distribution shifts.

\section{Conclusions}
\outlinebullets{
\item the main empirical lesson that absolute accuracy alone is insufficient;
\item the methodological contribution of excess properties as an evaluation lens;
\item the role of the ideal-mixture reference baseline and metrics for non-ideal mixture behavior in the proposed evaluation;
\item what the curated pure-plus-mixture datasets enable;
\item empirical guidance on evaluation settings for molecular mixture modeling;
\item the remaining open challenge for molecular mixture machine learning. 
}
\noindent
This work presents an evaluation framework for molecular mixture property prediction centered on mixing effects rather than solely on absolute accuracy. By combining curated pure and mixture datasets, leakage-aware split protocols, ideal-mixture baselines, and excess-property metrics, the benchmark makes it possible to separate interpolation between pure-component values from recovery of non-ideal intermolecular effects.

Across the evaluated tasks, the results show that strong absolute performance can mask weak recovery of non-ideal mixture behavior. This is most visible when models are compared under excess metrics and under strict split families. In particular, the comparison of molecule and pure-to-mixture splits expose a substantial remaining gap in transfer to mixtures built from unseen molecules, indicating that current models are often better at interpolation of pure properties than at learning the non-ideality of mixture behavior. The comparisons also provide practical guidance: MolT5 and DMPNN neural models are the strongest overall choice on absolute and excess accuracy, explicit interaction layers are not consistently beneficial, and weighted-sum aggregation is the most reliable default across both standard and stricter settings.

These findings suggest that progress in molecular mixture ML is limited by evaluation: without rigorous benchmarks, architectural advances are neither measurable nor reliable. Benchmarks that rely only on absolute prediction error can overstate model quality, especially when test mixtures remain close to seen chemistry. We therefore advocate reporting metrics for non-ideal mixture behavior alongside standard metrics, using the ideal-mixture reference baseline, and evaluating under split protocols that distinguish interpolation from genuine transfer. We hope this framework provides a reproducible foundation for that shift and helps establish a stronger standard for future molecular mixture benchmarks.

\section{Limitations}
\outlinebullets{
\item the chemical scope of the evaluation;
\item the strongest coverage in binary and low-order small-molecule mixtures;
\item dataset bias, measurement noise, and heterogeneous source conditions;
\item limits on how broadly the results can be generalized beyond the included datasets and properties;
\item the absence of mechanistic interpretability and full thermodynamic consistency guarantees;
\item the distinction between dataset limits and protocol limits.
}
This evaluation focuses on small-molecule mixtures and primarily covers binary and low-order systems, limiting direct applicability to high-order mixtures, polymers, or reactive mixtures. As the evaluation aggregates publicly available datasets, it inherits dataset-specific biases, measurement noise, and heterogeneity in experimental conditions. The empirical comparisons are also limited to the evaluated model families, so the resulting performance trends should be interpreted as benchmark findings for the tested methods rather than for all possible mixture-model architectures. While the proposed evaluation protocols assess mixture behavior trends and representation structure, they do not provide mechanistic or causal interpretability, nor do they enforce full thermodynamic consistency.

Potential positive impacts include reduced waste via more data-efficient screening of solvent and fuel mixtures. A potential negative impact is that improved model architectures derived from this benchmark could be applied to hazardous compound discovery.

\begin{ack}
J.G.R.\ acknowledges funding by the Deutsche Forschungsgemeinschaft (DFG, German Research Foundation) -- 466417970 -- within the Priority Programme ``SPP 2331: Machine Learning in Chemical Engineering''.
R.J.L.\ and F.H.V.\ acknowledge the Fonds Wetenschappelijk Onderzoek (FWO) for funding (G021924N). Resources and services used in this work were provided by the VSC (Flemish Supercomputer Center), funded by the Research Foundation -- Flanders (FWO) and the Flemish Government.
N.K.M.\ and W.H.G.\ acknowledge the Machine Learning for Pharmaceutical Discovery and Synthesis Consortium (MLPDS) for funding.
The authors declare no competing interests.
\end{ack}
{
\small
\bibliographystyle{unsrtnat_noid} 
\bibliography{refs}          
}


\appendix

\section{Technical Appendices and Supplementary Material}
\outlinebullets{
\item dataset standardization and filtering procedures;
\item training, hyperparameter, and reproducibility details omitted from the main text;
\item the detailed curation rules omitted from the main text for space;
\item the included property families and their scientific context;
\item the supporting tables and figures referenced from the main text;
\item the correspondence between raw sources and standardized outputs;
\item the reproducibility material for the evaluation artifacts described in the paper.
}

This appendix collects the implementation and data details that support the main evaluation. It first describes the shared curation and standardization pipeline, then summarizes the included datasets and their scientific scope, and finally reports the training and hyperparameter procedure used in the model comparisons. Additional result tables and figures referenced from the main text are collected in Section~\ref{app:additional-results}.

\FloatBarrier
\subsection{Data curation and standardization}
\label{app:data-curation}
\outlinebullets{
\item the preprocessing pipeline used to standardize all raw datasets;
\item the common validity and filtering rules applied across all tasks;
\item the common molecule and composition filtering rules applied across tasks;
\item the normalization of fractions, units, temperatures, and canonical identifiers;
\item the dataset-specific curation details for each property family;
\item the rejection and audit outputs that preserve preprocessing traceability.
}

All datasets were curated using dataset-specific standardization scripts (\texttt{datasets/*/standardize\_*.py}), with shared utilities (\texttt{datasets/utils/chem\_utils.py}, \texttt{datasets/utils/standardization\_utils.py}). Each script wrote (i) standardized processed CSV files and (ii) rejection CSV files containing raw fields and explicit rejection reasons, enabling traceable preprocessing.

\paragraph{Common rules across datasets}
Molecular identifiers (SMILES and/or InChI) were standardized with RDKit and accepted molecules were converted to a canonical explicit-hydrogen InChI representation. Entries were rejected if identifiers were missing or invalid, contained disconnected fragments (including dot-disconnected strings), carried non-zero formal charge, or contained metal atoms. For mixtures, component-fraction pairs were canonicalized by removing zero-fraction components, sorting components deterministically by standardized InChI, and merging duplicate components. Input fractions had to define a valid composition (up to a small numerical tolerance); exact percentage-style inputs (sum \(\approx 100\)) were first converted to fraction scale. Fractions were then rounded to three decimal places, with residual closure applied to the last component to enforce an exact sum of one after rounding. All mixture compositions were stored as molar fractions. Across all outputs, duplicate rows with identical canonical keys (composition, unit, and temperature where applicable) were collapsed by averaging the target value.

\paragraph{Flashpoint, DCN, and MON}
Flashpoint and DCN pure-compound data were read from \texttt{PureFuels.xlsx}; mixture data were read from \texttt{DatasetsFuelProperties.xlsx} (training/test sheets). Wide-format fuel mixture rows were parsed with a shared parser that inferred the trailing molar fraction by closure and then canonicalized component order and fractions. MON data were read from \texttt{published\_MONdata.csv} using indexed component and ``Mole fraction of cmp$i$'' columns, and split into pure versus mixture subsets based on the number of positive-fraction components after parsing. In the indexed parser, at most one missing component fraction was inferred as the remaining fraction to one; rows with two or more missing component fractions were rejected. DCN and MON units are dimensionless, while flashpoint values are stored in Celsius.

\paragraph{Viscosity}
Viscosity data were assembled from NIST (\texttt{nist\_vis/published\_logV.csv}), \texttt{PureFuels.xlsx}, and \texttt{DatasetsFuelProperties.xlsx}. NIST \texttt{logV} values were interpreted as base-10 logarithms and converted to natural-log viscosity by multiplying by \(\ln(10)\), while viscosity values from the other sources were interpreted on a linear scale and converted to \(\ln(\eta)\) only when positive. The standardized target unit is \(\ln(\mathrm{mPa}\!\cdot\!\mathrm{s})\). Temperatures were retained in Kelvin when provided (NIST) and set to 298~K for sources reported at ambient conditions.

\paragraph{Vaporization enthalpy}
The miscible solvent dataset (\texttt{MiscibleSolventData.csv}) was parsed from up to five component slots (\texttt{SMILES\_i}, \texttt{comp\_i}). Heat of vaporization was retained in \(\mathrm{kcal/mol}\). Although the source also reports enthalpy of mixing, it was excluded from the benchmark because it is an excess quantity derived from the same underlying property and would duplicate the evaluation.

\paragraph{Solvation free energy}
Solvation free-energy data were read from \texttt{MixSolvGH-QM.csv}. Solvent fractions were parsed from \texttt{frac\_solvent$i$} columns, with the final fraction inferred by closure, then canonicalized as described above. Pure and mixture outputs were generated from the same source.

\paragraph{Solubility}
Solubility data were curated from \texttt{BigSolDBv2\_1.csv} (pure) and \texttt{MixtureSolDB.csv} (pure and binary mixtures). The target was standardized as \(\log_{10}(S)\) on a mole-fraction basis: when the field \texttt{LogS(mole\_fraction)} was available it was used directly; otherwise, \(\log_{10}(\texttt{Solubility(mole\_fraction)})\) was computed for valid mole fractions (\(0 < x \le 1\)). Temperatures were stored in Kelvin. For \texttt{MixtureSolDB}, solvent composition could be reported as either mass fraction or mole fraction. When mass fractions were provided, solvent compositions were converted to mole fractions using RDKit molecular weights computed from standardized solvent InChIs.

\FloatBarrier
\subsection{Datasets overview}
\outlinebullets{
\item the scientific scope and interpretation of the included property families;
\item how the dataset collection spans multiple property regimes relevant to the evaluation.
}

\begin{figure}[h]
    \centering
    \includegraphics[width=1.0\linewidth]{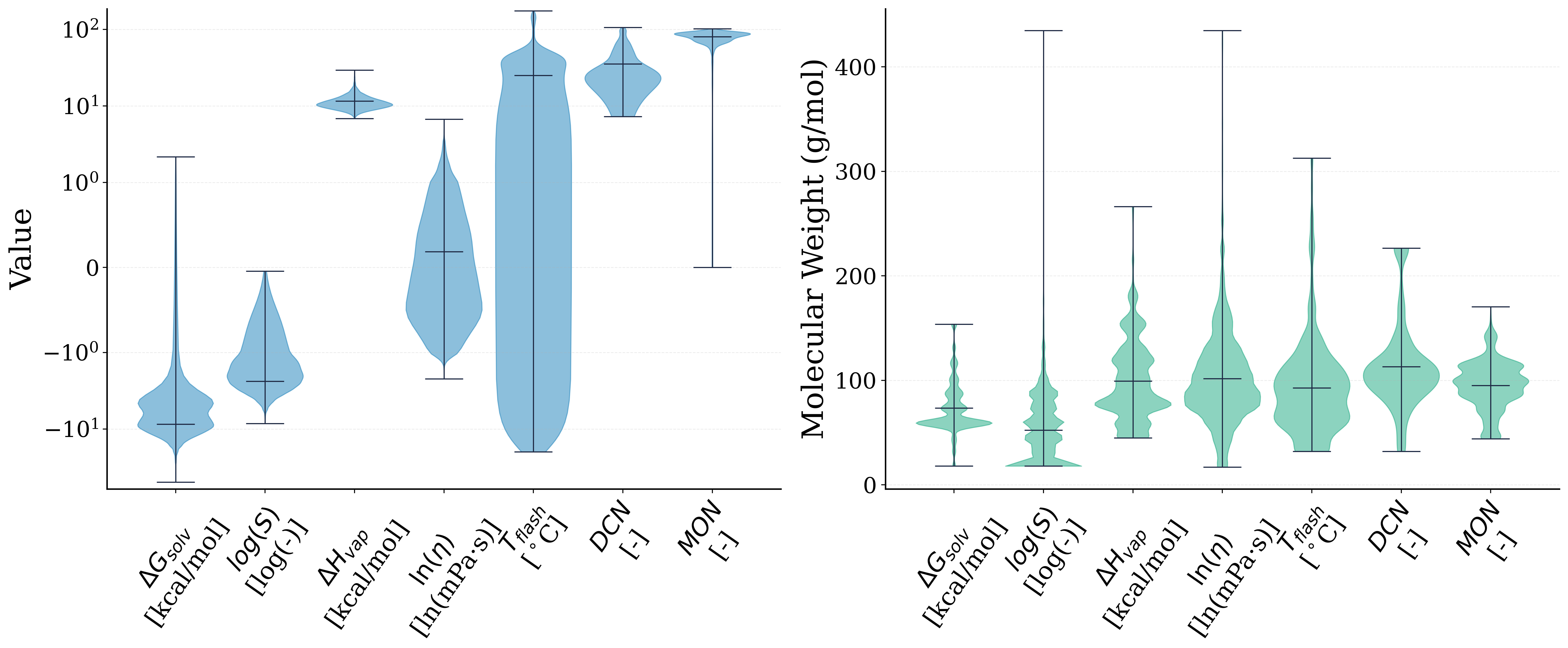}
    \caption{Distribution of mixture-property values and molecular weights across evaluation tasks.}
    \label{fig:mix-value-mw-dist}
\end{figure}

\begin{figure}[h]
    \centering
    \includegraphics[width=1.0\linewidth]{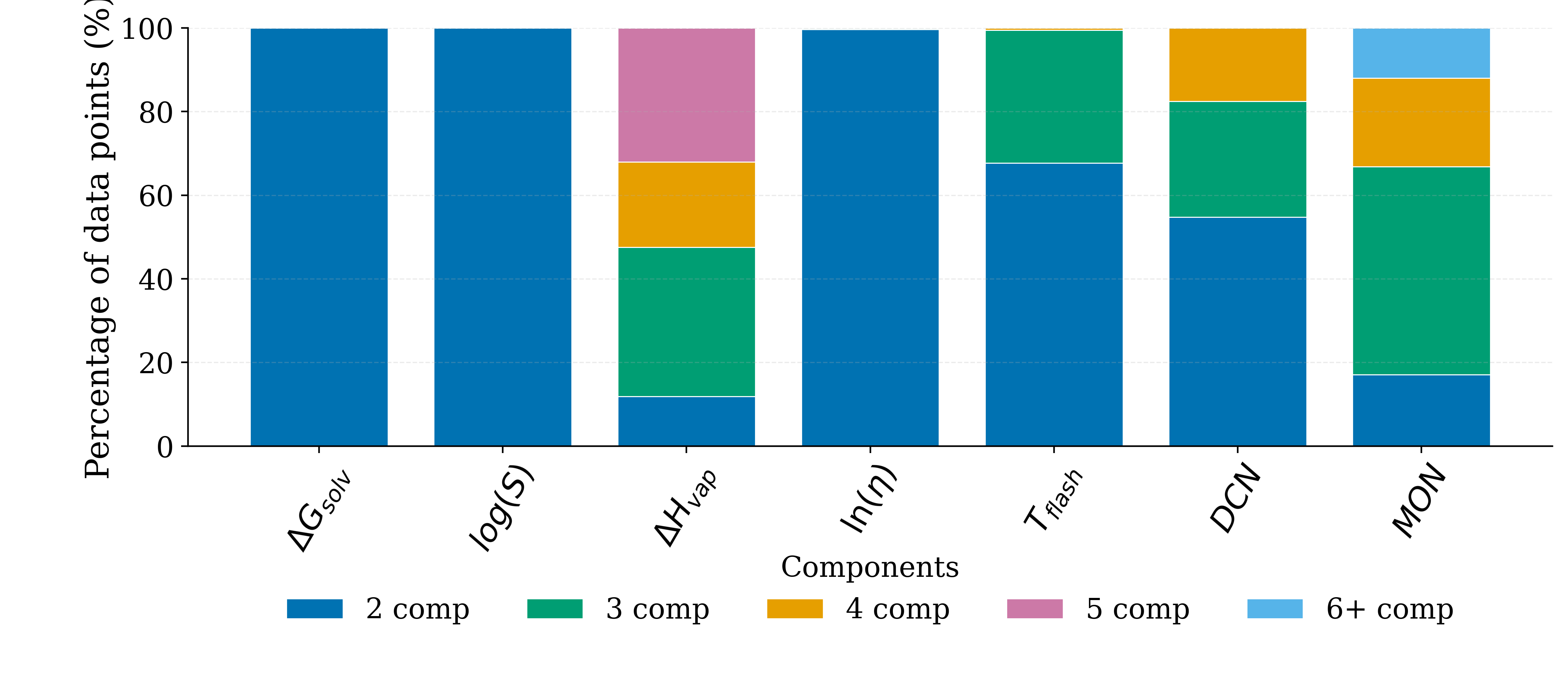}
    \caption{Distribution of number of components per mixture data point across evaluation datasets (6+ grouped).}
    \label{fig:component-percentage}
\end{figure}

\begin{figure}[h]
    \centering
    \includegraphics[width=1.0\linewidth]{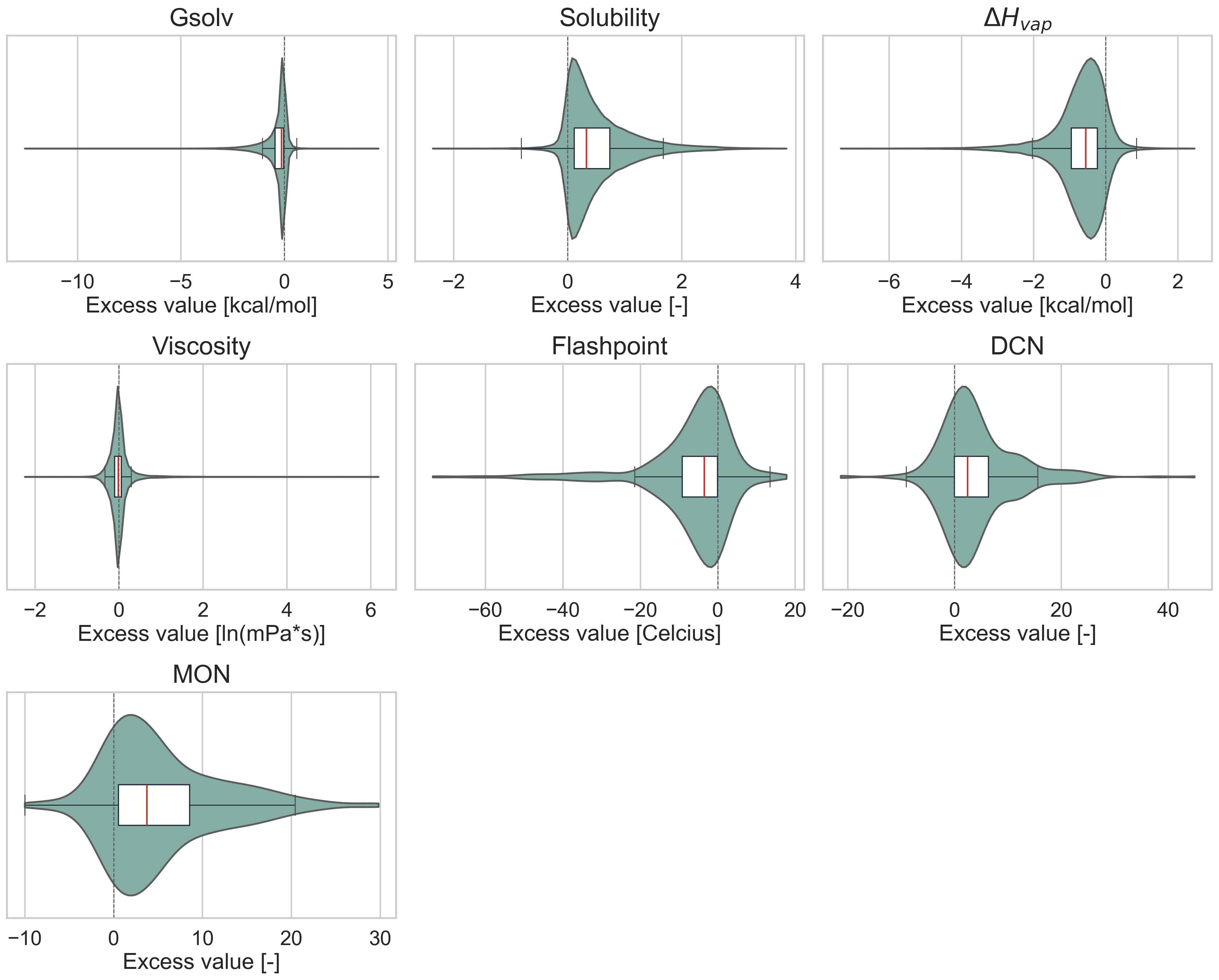}
    \caption{Distribution of excess-property values across mixture datasets, where excess values are defined relative to the ideal-mixture reference baseline used in the main text.}
    \label{fig:excess-value-violin-distributions}
\end{figure}

This section summarizes the property families covered by the paired pure and mixture datasets. Distributions of property values, molecular weights, mixture component counts, and excess-property values are shown in Figures~\ref{fig:mix-value-mw-dist}, \ref{fig:component-percentage}, and~\ref{fig:excess-value-violin-distributions}.

\paragraph{Solvation free energy}
Solvation free energy (\(\Delta G_{solv}\)) measures the Gibbs free energy change when transferring a solute from an ideal gas phase to a solvent. Computational datasets from published sources \cite{VERMEIRE2021129307, LEENHOUTS2025162232} were included, spanning pure, and binary solvent systems.
\paragraph{Solubility}
Solubility (\(\log(S)\)) describes the equilibrium concentration of a solute that can be dissolved in a solvent or solvent mixture at given conditions. Solubility datasets for organic compounds in mono- and binary-solvent systems were included from recent large-scale curations \cite{Krasnov_Malikov_Kiseleva_Tatarin_Bezzubov_2025, Malikov_Krasnov_Kiseleva_Meshcheriakova_Kuznetsov_Elistratov_Vasiyarov_Tatarin_Bezzubov_2025}.
\paragraph{Vaporization enthalpy}
Vaporization enthalpy (\(\Delta H_{vap}\)) measures the heat required to vaporize part of a liquid mixture. Molecular-dynamics data from Chew et al.\ \cite{Miscible_solvents} were included, providing a relatively low-noise computational setting for architecture evaluation. Enthalpy of mixing was not reported separately because it is an excess quantity derived from the same source and would duplicate the benchmark.
\paragraph{Viscosity}
Dynamic viscosity (\(\ln(\eta)\)) characterizes a fluid’s resistance to flow and shear deformation. Both pure and mixture viscosity datasets were compiled from published sources \cite{BILODEAU2023142454, LEENHOUTS2025133218, Larsson_Vermeire_Verhelst_2023}.
\paragraph{Flashpoint}
Flashpoint (\(T_{flash}\)) is the lowest temperature at which a liquid mixture produces sufficient vapor to ignite in the presence of an ignition source, making it a practically relevant safety and handling property for fuels and solvents. Experimental flashpoint datasets for both pure compounds and mixtures were included \cite{LEENHOUTS2025133218, Larsson_Vermeire_Verhelst_2023}, providing a low-data property where composition-dependent non-ideal effects can strongly affect risk-relevant behavior.
\paragraph{Derived cetane number and motor octane number}
The derived cetane number (DCN) and motor octane number (MON) are fuel performance indicators that quantify ignition quality in compression-ignition and knock resistance in spark-ignition engines, respectively. Experimental datasets containing both pure compounds and mixtures were compiled \cite{Larsson_Vermeire_Verhelst_2023, LEENHOUTS2025133218, Kuzhagaliyeva_Horváth_Williams_Nicolle_Sarathy_2022}.

\begin{table}[h]
\centering
\small
\caption{Summary of pure datasets and molecule-type coverage over corresponding mixture datasets.}
\label{tab:dataset-summary-pure}
\setlength{\tabcolsep}{4pt}
\resizebox{\textwidth}{!}{%
\begin{tabular}{lccrrcl}
\hline
Task & Type & Data points & \# Unique mol & Unit & \% Pure coverage & Source \\
\hline
$\Delta G_{solv}$ & Comp & 512846 & 285 & kcal/mol & 38.6\% & \cite{VERMEIRE2021129307} \\
$\Delta H_{vap}$ & Comp & 81 & 81 & kcal/mol & 100.0\% & \cite{Miscible_solvents} \\
$log(S)$ & Exp & 101564 & 218 & - & 75.3\% & \cite{Krasnov_Malikov_Kiseleva_Tatarin_Bezzubov_2025} \\
$\ln(\eta)$ & Exp & 3027 & 1389 & ln(mPa*s) & 88.1\% & \cite{BILODEAU2023142454, Larsson_Vermeire_Verhelst_2023} \\
$T_{flash}$ & Exp & 585 & 585 & $^{\circ}$C & 96.7\% & \cite{Larsson_Vermeire_Verhelst_2023} \\
$DCN$ & Exp & 229 & 229 & - & 74.4\% & \cite{Larsson_Vermeire_Verhelst_2023} \\
$MON$ & Exp & 332 & 332 & - & 73.6\% & \cite{Kuzhagaliyeva_Horváth_Williams_Nicolle_Sarathy_2022} \\
\hline
\end{tabular}%
}
\end{table}

\FloatBarrier
\subsection{Asset licenses}

For re-packaged dataset assets, we report the upstream source-specific licenses rather than a single coarse label per benchmark family:

\begin{itemize}
    \item $\Delta G_{solv}$: \texttt{MixSolvGH-QM.csv} (pure and mixture computational data), CC BY 4.0.
    \item $\Delta H_{vap}$: \texttt{MiscibleSolventData.csv}, CC BY-NC 4.0.
    \item $\log(S)$: \texttt{BigSolDBv2\_1.csv}, CC BY 4.0; \texttt{MixtureSolDB.csv}, CC BY 4.0.
    \item $\ln(\eta)$: \texttt{nist\_vis/published\_logV.csv} and \texttt{nist\_vis/compounds.csv}, CC BY 4.0; \texttt{PureFuels.xlsx} and \texttt{DatasetsFuelProperties.xlsx}, MIT.
    \item $T_{\mathrm{flash}}$: \texttt{PureFuels.xlsx} and \texttt{DatasetsFuelProperties.xlsx}, MIT.
    \item DCN: \texttt{PureFuels.xlsx} and \texttt{DatasetsFuelProperties.xlsx}, MIT.
    \item MON: \texttt{published\_MONdata.csv}, CC BY 4.0.
\end{itemize}

\noindent \textbf{Codebase:} MIT

\FloatBarrier
\subsection{Training Procedure}
\label{app:training-procedure}

All model families are trained on the same predefined 5-fold cross-validation splits for each dataset and split family, so fold-to-fold variability reflects the data partition rather than differences in protocol.

\paragraph{Optimization.}
All neural models (DMPNN\,+\,FFN and MolT5\,+\,FFN) are trained with the Adam optimizer and a Noam-like learning-rate schedule consisting of a linear warm-up over the first two epochs to a peak learning rate, followed by exponential decay to a final value. The default learning rate range is $10^{-4}$ (initial and final) to $10^{-3}$ (peak). The loss function is mean squared error (MSE) applied to standardized targets; target normalization uses a z-score scaler fitted on the training fold. All activations in message-passing layers, the predictor MLP, and the DeepSets sub-networks use LeakyReLU.

\paragraph{DMPNN\,+\,FFN defaults.}
The directed message-passing network uses a message hidden dimension of 101, a depth of 4 layers, and is followed by a 4-layer predictor MLP with hidden dimension 200. Mixtures are represented using weighted-sum aggregation over component embeddings by default. Models are trained for 50 epochs with a batch size of 8.

\paragraph{MolT5\,+\,FFN defaults.}
MolT5 component embeddings (256-dimensional) are passed through a 4-layer MLP with hidden dimension 200. Models are trained for 50 epochs with a batch size of 4.

\paragraph{RDKit\,+\,XGBoost defaults.}
Composition-aggregated 200-dimensional normalized RDKit descriptors are fitted with XGBoost (2000 estimators, max depth 5, learning rate 0.01, subsample 0.8). Early stopping uses 100 rounds monitored on the validation fold.

\paragraph{Compute resources.}
All reported models were trained on a single CPU-only node. The node configuration comprises 2 Intel Xeon Platinum 8468 (Sapphire Rapids) CPUs with 48 cores each (4 NUMA domains and 1 L3 cache per CPU), base and maximum frequencies of 2.1 and 3.8\,GHz, 256\,GiB RAM, 2500\,MiB default memory per core, and a 960\,GB local SSD. Within this node type, single benchmark runs were allocated 8--16 CPU cores and 32--64\,GB RAM for DMPNN\,+\,FFN, and 16--24 CPU cores and 64--96\,GB RAM for feature-based models (MolT5\,+\,FFN and RDKit\,+\,XGBoost), with wall-time limits of 8\,h and 12\,h respectively.

\FloatBarrier
\subsection{Hyperparameter Optimization}
\label{app:training-hparams}

\paragraph{Search algorithm.}
Hyperparameter search uses the Asynchronous Successive Halving Algorithm (ASHA)\cite{ASHA} implemented via Ray Tune\cite{liaw2018tune}, with a grace period of 3 epochs and a reduction factor of 2. The number of trials is 50 per configuration. The primary selection metric is validation RMSE.

\paragraph{Cross-dataset runs.}
In the round-2 manifest stage, a shared hyperparameter configuration is optimized to perform well across seven fold-00 mixture tasks: DCN, $\Delta H_{vap}$, flashpoint, $\Delta G_{solv}$ with the 100k computational training subset, MON, solubility, and viscosity. The purpose of these cross-dataset runs is to identify a robust reference setting that remains competitive across the physicochemical property datasets within the scope of this benchmark, rather than a separate best setting for each individual task. Each trial evaluates one configuration on all seven tasks, and the task-wise validation RMSE values are normalized with interquartile-range (IQR) scaling before averaging into a single aggregated objective. The configuration with the lowest aggregated normalized validation RMSE is selected as the shared reference setting.

\paragraph{Single-dataset runs.}
In the single-dataset stage, each model family is tuned independently for the corresponding fold-00 dataset task. This stage screens the broader round-2 search space (batch size, hidden dimensions, number of layers, learning rate, aggregation function, and interaction type; see Table~\ref{tab:hpo-search-space-model-overview}) and identifies competitive per-dataset configurations. For the neural mixture models, this stage re-opens the mixture-construction choices, so aggregation and interaction modules are not fixed in advance during tuning.

\paragraph{Validating the two-stage procedure.}
To confirm that the cross-dataset selection does not materially alter downstream conclusions relative to per-dataset tuning, Tables~\ref{tab:manifest-vs-single-rdkit}, \ref{tab:manifest-vs-single-molt5}, and~\ref{tab:manifest-vs-single-gnn} compare held-out test RMSE between the matched single-dataset and cross-dataset configurations for RDKit\,+\,XGBoost, MolT5\,+\,FFN, and DMPNN\,+\,FFN. Because the single-dataset searches also allow the interaction and aggregation choices to vary, these comparisons check whether fixing the simpler shared manifest configuration hides large gains from those modules. For RDKit\,+\,XGBoost and MolT5\,+\,FFN, the rankings and performance trends remain broadly consistent across both modes, which supports the cross-dataset selection as a representative reference configuration and is consistent with the limited and unstable gains observed in the targeted interaction and aggregation comparisons in Section~\ref{sec:interaction-aggregation-results}. For DMPNN\,+\,FFN, the picture is more mixed: the shared manifest run remains competitive on viscosity and some fuel-property tasks, but does not uniformly match the individually tuned runs. In addition, the corresponding single-dataset solubility GNN HPO run is incomplete in the inspected round-2 artifact set, so that row is omitted from the direct comparison.

\paragraph{Applied benchmark defaults.}
The cross-dataset selections define the default-hyperparameter benchmark. For RDKit\,+\,XGBoost and MolT5\,+\,FFN, these manifest selections are applied directly by the current training script. For DMPNN\,+\,FFN, the benchmark default matches the shared manifest choice for aggregation, interaction setting, batch size, message width and depth, and peak learning rate, while retaining a 4-layer predictor MLP instead of the manifest-selected 3-layer predictor.

\begin{table*}[h]
\centering
\small
\caption{Hyperparameter search spaces. Bold marks the cross-dataset selection used as the training default.}
\label{tab:hpo-search-space-model-overview}
\setlength{\tabcolsep}{5pt}
\begin{tabular}{>{\raggedright\arraybackslash}p{0.21\textwidth}>{\raggedright\arraybackslash}p{0.23\textwidth}>{\raggedright\arraybackslash}p{0.23\textwidth}>{\raggedright\arraybackslash}p{0.23\textwidth}}
\toprule
Hyperparameter & GNN & MolT5 + FFN & RDKit + XGBoost \\
\midrule
Aggregation & \textbf{weightedsum}, cat, deepsets, attentive, set2set & concat, \textbf{weightedsum}, deepsets, attentive, set2set & concat, \textbf{weightedsum} \\
Interaction setting & \textbf{none}, molecular, interaction & \textbf{none}, self\_attention & -- \\
Batch size & 4, 16, 32, \textbf{64} & 4, \textbf{16}, 32, 64 & -- \\
Learning rate & loguniform\allowbreak$[0.000300, 0.00300]$, \textbf{0.00178} & loguniform\allowbreak$[0.000100, 0.00300]$, \textbf{0.000111} & loguniform\allowbreak$[0.0100, 0.200]$, \textbf{0.118} \\
Message hidden dim & 64, \textbf{128}, 192 & -- & -- \\
Message passing depth & 3, \textbf{4}, 5 & -- & -- \\
Predictor hidden dim & 100, \textbf{200}, 300 & -- & -- \\
Predictor layers & \textbf{3}, 4, 5 & -- & -- \\
FFN hidden dim & -- & 100, 200, 300, \textbf{400} & -- \\
FFN layers & -- & 3, \textbf{4}, 5 & -- \\
XGBoost estimators & -- & -- & 500, 1000, \textbf{2000}, 4000 \\
XGBoost max depth & -- & -- & \textbf{3}, 5, 7 \\
XGBoost subsample & -- & -- & \textbf{0.800}, 1.00 \\
XGBoost colsample\_bytree & -- & -- & \textbf{0.800}, 1.00 \\
XGBoost $\lambda$ & -- & -- & \textbf{0.500}, 1.00, 2.00 \\
\bottomrule
\end{tabular}
\end{table*}

\begin{table}[h]
\centering
\caption{Fold-00 test RMSE for RDKit + XGBoost under per-dataset versus cross-dataset hyperparameter optimization. Lower is better.}
\label{tab:manifest-vs-single-rdkit}
\small
\begin{tabular}{lrrrr}
\toprule
Dataset & Unit & Single RMSE & Cross-dataset RMSE & $\Delta$ \\
\midrule
DCN & [-] & 6.32 & 4.90 & -1.42 \\
$\Delta H_{vap}$ & [kcal/mol] & 0.582 & 0.563 & -0.0184 \\
$T_{flash}$ & [$^\circ$C] & 15.6 & 14.2 & -1.45 \\
$\Delta G_{solv}$ (100k train) & [kcal/mol] & 0.415 & 0.592 & +0.177 \\
MON & [-] & 6.64 & 7.27 & +0.630 \\
$\log(S)$ & [log(-)] & 0.429 & 0.471 & +0.0418 \\
$\ln(\eta)$ & [ln(mPa·s)] & 0.155 & 0.156 & +0.000361 \\
\bottomrule
\end{tabular}
\end{table}

\begin{table}[h]
\centering
\caption{Fold-00 test RMSE for MolT5 + FFN under per-dataset versus cross-dataset hyperparameter optimization. Lower is better.}
\label{tab:manifest-vs-single-molt5}
\small
\begin{tabular}{lrrrr}
\toprule
Dataset & Unit & Single RMSE & Cross-dataset RMSE & $\Delta$ \\
\midrule
DCN & [-] & 3.36 & 2.28 & -1.08 \\
$\Delta H_{vap}$ & [kcal/mol] & 0.218 & 0.313 & +0.0951 \\
$T_{flash}$ & [$^\circ$C] & 15.5 & 14.2 & -1.32 \\
$\Delta G_{solv}$ (100k train) & [kcal/mol] & 0.617 & 0.391 & -0.226 \\
MON & [-] & 1.95 & 2.56 & +0.610 \\
$\log(S)$ & [log(-)] & 0.428 & 0.434 & +0.00624 \\
$\ln(\eta)$ & [ln(mPa·s)] & 0.146 & 0.184 & +0.0377 \\
\bottomrule
\end{tabular}
\end{table}

\begin{table}[h]
\centering
\caption{Fold-00 test RMSE for DMPNN + FFN under per-dataset versus cross-dataset hyperparameter optimization. Lower is better.}
\label{tab:manifest-vs-single-gnn}
\small
\begin{tabular}{lrrrr}
\toprule
Dataset & Unit & Single RMSE & Cross-dataset RMSE & $\Delta$ \\
\midrule
DCN & [-] & 1.33 & 3.39 & +2.06 \\
$\Delta H_{vap}$ & [kcal/mol] & 0.131 & 0.179 & +0.0477 \\
$T_{flash}$ & [$^\circ$C] & 9.25 & 14.6 & +5.35 \\
$\Delta G_{solv}$ (100k train) & [kcal/mol] & 0.350 & 0.371 & +0.0204 \\
MON & [-] & 4.84 & 3.74 & -1.10 \\
$\ln(\eta)$ & [ln(mPa·s)] & 0.147 & 0.150 & +0.00306 \\
\bottomrule
\end{tabular}
\vspace{0.25em}\\{\footnotesize Omitted rows without matching completed single and manifest round-2 artifacts: $\log(S)$.}
\end{table}

\FloatBarrier
\subsection{Additional result tables and figures}
\label{app:additional-results}

\subsubsection{Split-comparison metrics}
\label{app:split-comparison-metrics}
\begin{figure}[h]
    \centering
    \includegraphics[width=0.72\linewidth]{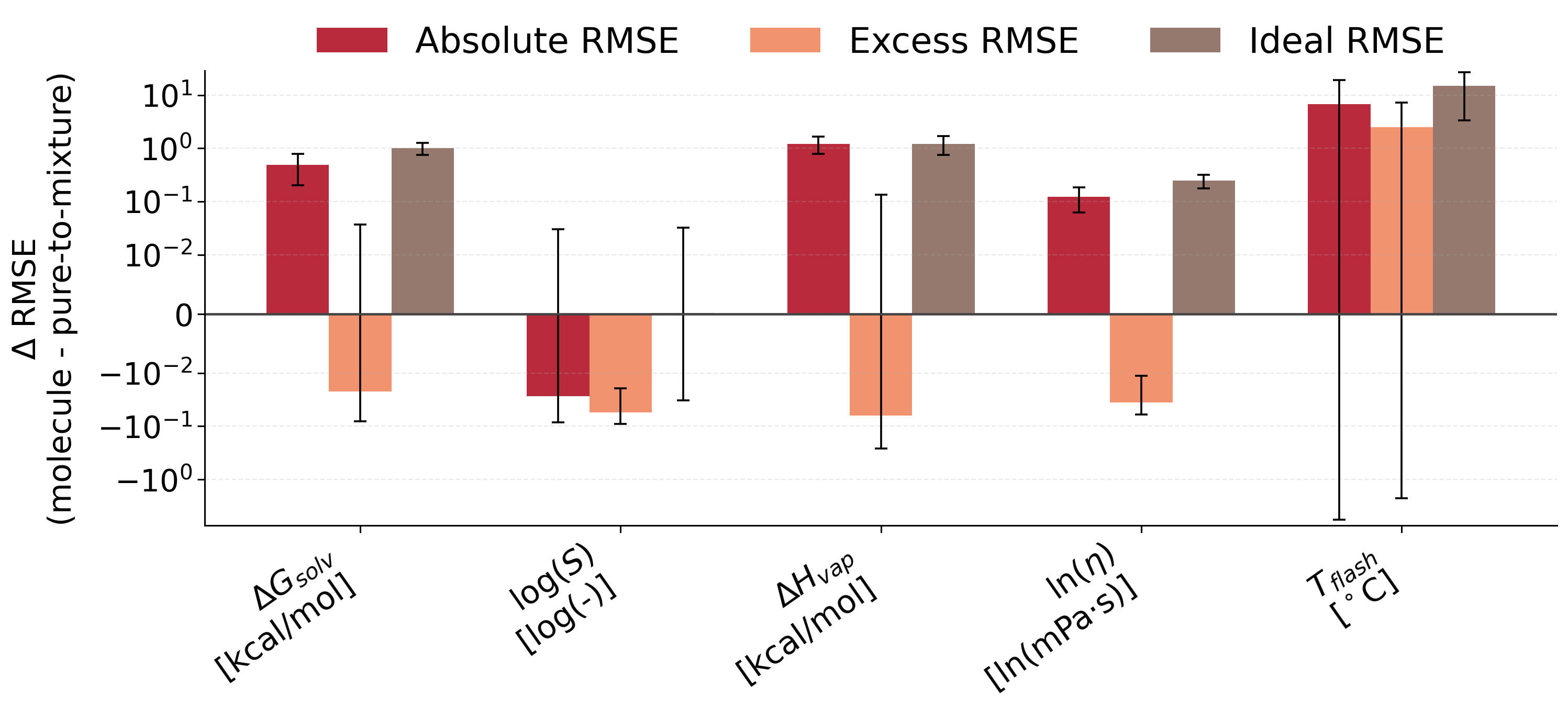}
    \caption{Differences in absolute, excess, and ideal RMSE from pure-to-mixture to molecule splits for the MolT5 + FFN model family.}
    \label{fig:matrix1-molt5-pure-to-mixture-vs-molecule-type-rmse}
\end{figure}

Table~\ref{tab:default-hparam-matrix1-excess-rmse-picture-style-full} extends the main-text excess RMSE table to all three splits, including pure-to-mixture. The following three tables report the corresponding absolute RMSE, ideal RMSE, and Kendall rank-correlation summaries across the same splits, providing the fuller metric picture behind the main-text discussion of mixture, molecule, and pure-to-mixture generalization. Figure~\ref{fig:matrix1-molt5-pure-to-mixture-vs-molecule-type-rmse} completes the split-comparison view with the MolT5\,+\,FFN model family, complementing the DMPNN\,+\,FFN and RDKit\,+\,XGBoost panels in the main text.

\begin{table*}[h]
\centering
\scriptsize
\caption{Excess RMSE across all data splits and datasets. Values are mean $\pm$ std over completed folds. Bold marks the best mean; underlining marks values within two standard deviations of the best mean.}
\label{tab:default-hparam-matrix1-excess-rmse-picture-style-full}
\setlength{\tabcolsep}{4pt}
\resizebox{\textwidth}{!}{%
\begin{tabular}{llccccccc}
\toprule
Split & Model type & \shortstack{$\Delta G_{solv}$\\ {[kcal/mol]}} & \shortstack{$\log(S)$\\ {[log(-)]}} & \shortstack{$\Delta H_{vap}$\\ {[kcal/mol]}} & \shortstack{$\ln(\eta)$\\ {[ln(mPa·s)]}} & \shortstack{$T_{flash}$\\ {[$^\circ$C]}} & \shortstack{DCN\\ {[-]}} & \shortstack{MON\\ {[-]}} \\
\midrule
Random & Ideal baseline & 0.71 $\pm$ 0.0064 & 0.77 $\pm$ 0.0073 & 1.0 $\pm$ 0.0082 & 0.30 $\pm$ 0.015 & 14. $\pm$ 0.78 & 8.2 $\pm$ 1.4 & 8.4 $\pm$ 0.74 \\
 & DMPNN + FFN & \underline{0.14 $\pm$ 0.0080} & \textbf{0.075 $\pm$ 0.0040} & \textbf{0.066 $\pm$ 0.0048} & \textbf{0.096 $\pm$ 0.0027} & 10. $\pm$ 0.82 & 5.3 $\pm$ 1.3 & \underline{3.9 $\pm$ 0.34} \\
 & MolT5 + FFN & \textbf{0.13 $\pm$ 0.0057} & 0.15 $\pm$ 0.049 & 0.14 $\pm$ 0.041 & 0.11 $\pm$ 0.0045 & 11. $\pm$ 0.67 & \underline{4.6 $\pm$ 0.94} & \underline{3.9 $\pm$ 0.60} \\
 & RDKit + XGBoost & 0.34 $\pm$ 0.0015 & 0.29 $\pm$ 0.0030 & 0.20 $\pm$ 0.0044 & 0.14 $\pm$ 0.0091 & \textbf{4.3 $\pm$ 0.44} & \textbf{3.7 $\pm$ 0.75} & \textbf{2.8 $\pm$ 0.62} \\
\midrule
Mixture & Ideal baseline & 0.71 $\pm$ 0.053 & 0.74 $\pm$ 0.22 & 1.1 $\pm$ 0.23 & 0.30 $\pm$ 0.035 & \underline{13. $\pm$ 5.4} & \underline{7.9 $\pm$ 3.3} & \underline{7.6 $\pm$ 3.6} \\
 & DMPNN + FFN & \textbf{0.18 $\pm$ 0.037} & \textbf{0.36 $\pm$ 0.062} & \textbf{0.11 $\pm$ 0.064} & \textbf{0.18 $\pm$ 0.034} & \underline{11. $\pm$ 4.7} & \underline{6.3 $\pm$ 3.2} & \underline{5.3 $\pm$ 1.0} \\
 & MolT5 + FFN & \underline{0.19 $\pm$ 0.034} & \underline{0.37 $\pm$ 0.042} & \underline{0.13 $\pm$ 0.070} & \underline{0.20 $\pm$ 0.042} & \underline{11. $\pm$ 4.7} & \textbf{5.0 $\pm$ 3.1} & \textbf{5.0 $\pm$ 1.7} \\
 & RDKit + XGBoost & 0.39 $\pm$ 0.026 & \underline{0.37 $\pm$ 0.057} & 0.29 $\pm$ 0.16 & \underline{0.21 $\pm$ 0.048} & \textbf{9.1 $\pm$ 3.2} & \underline{5.9 $\pm$ 1.7} & \underline{5.5 $\pm$ 1.8} \\
\midrule
Molecule & Ideal baseline & \underline{0.65 $\pm$ 0.23} & \underline{0.81 $\pm$ 0.39} & 0.87 $\pm$ 0.33 & \underline{0.21 $\pm$ 0.062} & \underline{13. $\pm$ 8.3} & -- & -- \\
 & DMPNN + FFN & \textbf{0.52 $\pm$ 0.17} & \underline{0.74 $\pm$ 0.33} & \textbf{0.50 $\pm$ 0.15} & \underline{0.24 $\pm$ 0.064} & \textbf{13. $\pm$ 6.1} & -- & -- \\
 & MolT5 + FFN & \underline{0.59 $\pm$ 0.20} & \underline{0.80 $\pm$ 0.37} & \underline{0.57 $\pm$ 0.18} & \textbf{0.21 $\pm$ 0.061} & \underline{15. $\pm$ 7.1} & -- & -- \\
 & RDKit + XGBoost & \underline{0.60 $\pm$ 0.14} & \textbf{0.68 $\pm$ 0.32} & 0.80 $\pm$ 0.20 & \underline{0.26 $\pm$ 0.050} & \underline{17. $\pm$ 6.1} & -- & -- \\
\midrule
Pure-to-mixture & DMPNN + FFN & \textbf{0.51 $\pm$ 0.17} & \underline{0.83 $\pm$ 0.39} & \underline{0.96 $\pm$ 0.29} & \textbf{0.23 $\pm$ 0.057} & \underline{13. $\pm$ 7.4} & -- & -- \\
 & MolT5 + FFN & \underline{0.61 $\pm$ 0.20} & \underline{0.85 $\pm$ 0.39} & \textbf{0.63 $\pm$ 0.30} & \underline{0.24 $\pm$ 0.067} & \textbf{13. $\pm$ 8.9} & -- & -- \\
 & RDKit + XGBoost & \underline{0.71 $\pm$ 0.18} & \textbf{0.76 $\pm$ 0.31} & 1.7 $\pm$ 0.48 & 0.55 $\pm$ 0.076 & \underline{16. $\pm$ 4.0} & -- & -- \\
\bottomrule
\end{tabular}%
}
\end{table*}

\begin{table*}[t!]
\centering
\scriptsize
\caption{Absolute RMSE across data splits. Values are mean $\pm$ std over completed folds. Bold marks the best mean; underlining marks values within two standard deviations of the best mean.}
\label{tab:default-hparam-matrix1-absolute-rmse-picture-style}
\setlength{\tabcolsep}{4pt}
\resizebox{\textwidth}{!}{%
\begin{tabular}{llccccc}
\toprule
Split & Model type & \shortstack{$\Delta G_{solv}$\\ {[kcal/mol]}} & \shortstack{$\log(S)$\\ {[log(-)]}} & \shortstack{$\Delta H_{vap}$\\ {[kcal/mol]}} & \shortstack{$\ln(\eta)$\\ {[ln(mPa·s)]}} & \shortstack{$T_{flash}$\\ {[$^\circ$C]}} \\
\midrule
Random & Ideal baseline & 0.71 $\pm$ 0.0064 & 0.77 $\pm$ 0.0073 & 1.0 $\pm$ 0.0082 & 0.30 $\pm$ 0.015 & 14. $\pm$ 0.78 \\
 & DMPNN + FFN & 0.25 $\pm$ 0.0065 & \textbf{0.066 $\pm$ 0.0013} & \textbf{0.072 $\pm$ 0.014} & \textbf{0.10 $\pm$ 0.0056} & 6.3 $\pm$ 0.33 \\
 & MolT5 + FFN & \textbf{0.20 $\pm$ 0.020} & 0.12 $\pm$ 0.022 & 0.11 $\pm$ 0.040 & 0.12 $\pm$ 0.0074 & 6.0 $\pm$ 0.27 \\
 & RDKit + XGBoost & 0.60 $\pm$ 0.0041 & 0.25 $\pm$ 0.0015 & 0.19 $\pm$ 0.0043 & 0.14 $\pm$ 0.010 & \textbf{4.3 $\pm$ 0.45} \\
\midrule
Mixture & Ideal baseline & 0.71 $\pm$ 0.053 & 0.74 $\pm$ 0.22 & 1.1 $\pm$ 0.23 & 0.30 $\pm$ 0.035 & \underline{13. $\pm$ 5.4} \\
 & DMPNN + FFN & \underline{0.28 $\pm$ 0.085} & \textbf{0.38 $\pm$ 0.060} & \textbf{0.11 $\pm$ 0.052} & \textbf{0.20 $\pm$ 0.045} & \underline{9.4 $\pm$ 5.0} \\
 & MolT5 + FFN & \textbf{0.26 $\pm$ 0.054} & \underline{0.41 $\pm$ 0.034} & \underline{0.13 $\pm$ 0.074} & \underline{0.22 $\pm$ 0.048} & \textbf{9.0 $\pm$ 3.1} \\
 & RDKit + XGBoost & 0.65 $\pm$ 0.060 & \underline{0.45 $\pm$ 0.043} & 0.29 $\pm$ 0.16 & \underline{0.21 $\pm$ 0.060} & \underline{9.1 $\pm$ 3.2} \\
\midrule
Molecule & Ideal baseline & \textbf{0.65 $\pm$ 0.23} & \underline{0.81 $\pm$ 0.39} & \textbf{0.87 $\pm$ 0.33} & \textbf{0.21 $\pm$ 0.062} & \textbf{13. $\pm$ 8.3} \\
 & DMPNN + FFN & \underline{1.1 $\pm$ 0.35} & \underline{0.72 $\pm$ 0.32} & \underline{1.4 $\pm$ 0.23} & 0.36 $\pm$ 0.11 & \underline{14. $\pm$ 4.3} \\
 & MolT5 + FFN & 1.2 $\pm$ 0.22 & \underline{0.80 $\pm$ 0.37} & 2.6 $\pm$ 0.57 & 0.38 $\pm$ 0.065 & \underline{22. $\pm$ 9.7} \\
 & RDKit + XGBoost & \underline{1.1 $\pm$ 0.12} & \textbf{0.70 $\pm$ 0.24} & 1.6 $\pm$ 0.27 & \underline{0.31 $\pm$ 0.052} & \underline{18. $\pm$ 4.5} \\
\midrule
Pure-to-mixture & DMPNN + FFN & \textbf{0.64 $\pm$ 0.20} & \underline{0.81 $\pm$ 0.37} & 3.3 $\pm$ 0.97 & \textbf{0.24 $\pm$ 0.057} & \textbf{14. $\pm$ 5.7} \\
 & MolT5 + FFN & \underline{0.68 $\pm$ 0.21} & \underline{0.84 $\pm$ 0.37} & \textbf{1.3 $\pm$ 0.34} & \underline{0.26 $\pm$ 0.070} & \underline{15. $\pm$ 5.0} \\
 & RDKit + XGBoost & \underline{1.0 $\pm$ 0.13} & \textbf{0.77 $\pm$ 0.25} & \underline{1.6 $\pm$ 0.47} & 0.54 $\pm$ 0.083 & \underline{15. $\pm$ 4.3} \\
\bottomrule
\end{tabular}%
}
\end{table*}

\begin{table*}[t!]
\centering
\scriptsize
\caption{Ideal RMSE across data splits. Values are mean $\pm$ std over completed folds. Bold marks the best mean; underlining marks values within two standard deviations of the best mean.}
\label{tab:default-hparam-matrix1-ideal-rmse-picture-style}
\setlength{\tabcolsep}{4pt}
\resizebox{\textwidth}{!}{%
\begin{tabular}{llccccc}
\toprule
Split & Model type & \shortstack{$\Delta G_{solv}$\\ {[kcal/mol]}} & \shortstack{$\log(S)$\\ {[log(-)]}} & \shortstack{$\Delta H_{vap}$\\ {[kcal/mol]}} & \shortstack{$\ln(\eta)$\\ {[ln(mPa·s)]}} & \shortstack{$T_{flash}$\\ {[$^\circ$C]}} \\
\midrule
Random & DMPNN + FFN & \underline{0.17 $\pm$ 0.0089} & \textbf{0.060 $\pm$ 0.0035} & \underline{0.052 $\pm$ 0.013} & \textbf{0.058 $\pm$ 0.0024} & 8.0 $\pm$ 0.54 \\
 & MolT5 + FFN & \textbf{0.16 $\pm$ 0.023} & 0.16 $\pm$ 0.080 & 0.16 $\pm$ 0.078 & 0.081 $\pm$ 0.0041 & 9.5 $\pm$ 1.2 \\
 & RDKit + XGBoost & 0.47 $\pm$ 0.0051 & 0.36 $\pm$ 0.013 & \textbf{0.051 $\pm$ 0.0034} & 0.14 $\pm$ 0.0070 & \textbf{0.20 $\pm$ 0.020} \\
\midrule
Mixture & DMPNN + FFN & \underline{0.17 $\pm$ 0.022} & \textbf{0.062 $\pm$ 0.010} & \underline{0.073 $\pm$ 0.046} & \textbf{0.060 $\pm$ 0.0075} & 6.3 $\pm$ 2.9 \\
 & MolT5 + FFN & \textbf{0.16 $\pm$ 0.018} & 0.13 $\pm$ 0.035 & 0.13 $\pm$ 0.083 & \underline{0.075 $\pm$ 0.012} & 7.0 $\pm$ 2.4 \\
 & RDKit + XGBoost & 0.48 $\pm$ 0.022 & 0.38 $\pm$ 0.041 & \textbf{0.048 $\pm$ 0.013} & 0.16 $\pm$ 0.041 & \textbf{0.17 $\pm$ 0.030} \\
\midrule
Molecule & DMPNN + FFN & \underline{1.3 $\pm$ 0.72} & \textbf{0.059 $\pm$ 0.025} & \textbf{1.3 $\pm$ 0.18} & 0.38 $\pm$ 0.17 & \underline{8.7 $\pm$ 3.2} \\
 & MolT5 + FFN & \underline{1.2 $\pm$ 0.23} & 0.13 $\pm$ 0.033 & 2.4 $\pm$ 0.55 & \underline{0.33 $\pm$ 0.067} & 25. $\pm$ 11. \\
 & RDKit + XGBoost & \textbf{1.0 $\pm$ 0.15} & 0.41 $\pm$ 0.057 & \underline{1.4 $\pm$ 0.27} & \textbf{0.27 $\pm$ 0.032} & \textbf{8.5 $\pm$ 2.5} \\
\midrule
Pure-to-mixture & DMPNN + FFN & 0.33 $\pm$ 0.093 & \textbf{0.074 $\pm$ 0.014} & 3.0 $\pm$ 1.1 & \textbf{0.061 $\pm$ 0.0042} & 6.0 $\pm$ 1.2 \\
 & MolT5 + FFN & \textbf{0.22 $\pm$ 0.020} & 0.12 $\pm$ 0.032 & \underline{1.1 $\pm$ 0.30} & 0.080 $\pm$ 0.026 & 10. $\pm$ 3.4 \\
 & RDKit + XGBoost & 0.59 $\pm$ 0.043 & 0.39 $\pm$ 0.062 & \textbf{0.90 $\pm$ 0.75} & 0.12 $\pm$ 0.012 & \textbf{2.8 $\pm$ 1.4} \\
\bottomrule
\end{tabular}%
}
\end{table*}

\begin{table*}[h]
\centering
\scriptsize
\caption{Absolute Kendall rank correlation across random, mixture, molecule, and pure-to-mixture splits, excluding DCN and MON. The ideal baseline uses ideal-mixture predictions. Values are mean $\pm$ std over completed folds. Bold marks the best mean per split; underlining marks values within two standard deviations of the best mean.}
\label{tab:default-hparam-matrix1-absolute-kendall-picture-style}
\setlength{\tabcolsep}{4pt}
\resizebox{\textwidth}{!}{%
\begin{tabular}{llccccc}
\toprule
Split & Model type & \shortstack{$\Delta G_{solv}$\\ {[kcal/mol]}} & \shortstack{$\Delta H_{vap}$\\ {[kcal/mol]}} & \shortstack{Solubility\\ {[log(-)]}} & \shortstack{Viscosity\\ {[ln(mPa·s)]}} & \shortstack{Flashpoint\\ {[$^\circ$C]}} \\
\midrule
Random & Ideal baseline & 0.93 $\pm$ 0.00092 & 0.83 $\pm$ 0.0036 & 0.76 $\pm$ 0.0037 & 0.84 $\pm$ 0.0018 & 0.76 $\pm$ 0.028 \\
 & DMPNN + FFN & \underline{0.98 $\pm$ 0.0019} & \textbf{0.98 $\pm$ 0.00056} & \textbf{0.98 $\pm$ 0.00037} & \textbf{0.94 $\pm$ 0.0018} & 0.85 $\pm$ 0.0079 \\
 & MolT5 + FFN & \textbf{0.98 $\pm$ 0.0024} & 0.98 $\pm$ 0.0046 & 0.95 $\pm$ 0.0052 & 0.93 $\pm$ 0.0015 & 0.88 $\pm$ 0.0090 \\
 & RDKit + XGBoost & 0.92 $\pm$ 0.00067 & 0.95 $\pm$ 0.00075 & 0.87 $\pm$ 0.0013 & 0.92 $\pm$ 0.0013 & \textbf{0.90 $\pm$ 0.0095} \\
\midrule
Mixture & Ideal baseline & 0.93 $\pm$ 0.0030 & 0.83 $\pm$ 0.041 & 0.76 $\pm$ 0.052 & 0.84 $\pm$ 0.013 & \textbf{0.75 $\pm$ 0.14} \\
 & DMPNN + FFN & \underline{0.97 $\pm$ 0.0063} & \textbf{0.98 $\pm$ 0.0043} & \textbf{0.81 $\pm$ 0.024} & \textbf{0.90 $\pm$ 0.013} & \underline{0.74 $\pm$ 0.17} \\
 & MolT5 + FFN & \textbf{0.97 $\pm$ 0.0035} & \underline{0.97 $\pm$ 0.0050} & \underline{0.80 $\pm$ 0.015} & \underline{0.88 $\pm$ 0.016} & \underline{0.74 $\pm$ 0.16} \\
 & RDKit + XGBoost & 0.92 $\pm$ 0.0056 & 0.93 $\pm$ 0.0094 & 0.75 $\pm$ 0.020 & \underline{0.88 $\pm$ 0.020} & \underline{0.71 $\pm$ 0.15} \\
\midrule
Molecule & Ideal baseline & \textbf{0.94 $\pm$ 0.024} & \textbf{0.83 $\pm$ 0.037} & \underline{0.73 $\pm$ 0.091} & \textbf{0.85 $\pm$ 0.025} & \textbf{0.68 $\pm$ 0.27} \\
 & DMPNN + FFN & 0.84 $\pm$ 0.046 & 0.68 $\pm$ 0.079 & \textbf{0.73 $\pm$ 0.076} & 0.75 $\pm$ 0.072 & \underline{0.56 $\pm$ 0.24} \\
 & MolT5 + FFN & 0.87 $\pm$ 0.024 & 0.53 $\pm$ 0.12 & \underline{0.70 $\pm$ 0.084} & 0.68 $\pm$ 0.064 & \underline{0.36 $\pm$ 0.39} \\
 & RDKit + XGBoost & 0.87 $\pm$ 0.012 & 0.66 $\pm$ 0.057 & \underline{0.66 $\pm$ 0.044} & 0.75 $\pm$ 0.033 & \underline{0.45 $\pm$ 0.32} \\
\midrule
Pure-to-mixture & DMPNN + FFN & \underline{0.94 $\pm$ 0.020} & \underline{0.50 $\pm$ 0.14} & \textbf{0.71 $\pm$ 0.082} & \textbf{0.83 $\pm$ 0.029} & \textbf{0.57 $\pm$ 0.25} \\
 & MolT5 + FFN & \textbf{0.94 $\pm$ 0.019} & \textbf{0.64 $\pm$ 0.14} & \underline{0.68 $\pm$ 0.087} & \underline{0.80 $\pm$ 0.048} & \underline{0.53 $\pm$ 0.25} \\
 & RDKit + XGBoost & 0.88 $\pm$ 0.012 & \underline{0.58 $\pm$ 0.062} & \underline{0.65 $\pm$ 0.049} & 0.54 $\pm$ 0.079 & \underline{0.53 $\pm$ 0.23} \\
\bottomrule
\end{tabular}%
}
\end{table*}

\subsubsection{Featurization versus ideal-mixture baseline: all datasets}
\label{app:featurization-vs-baseline-all}
\begin{figure}[h]
    \centering
    \includegraphics[width=\linewidth]{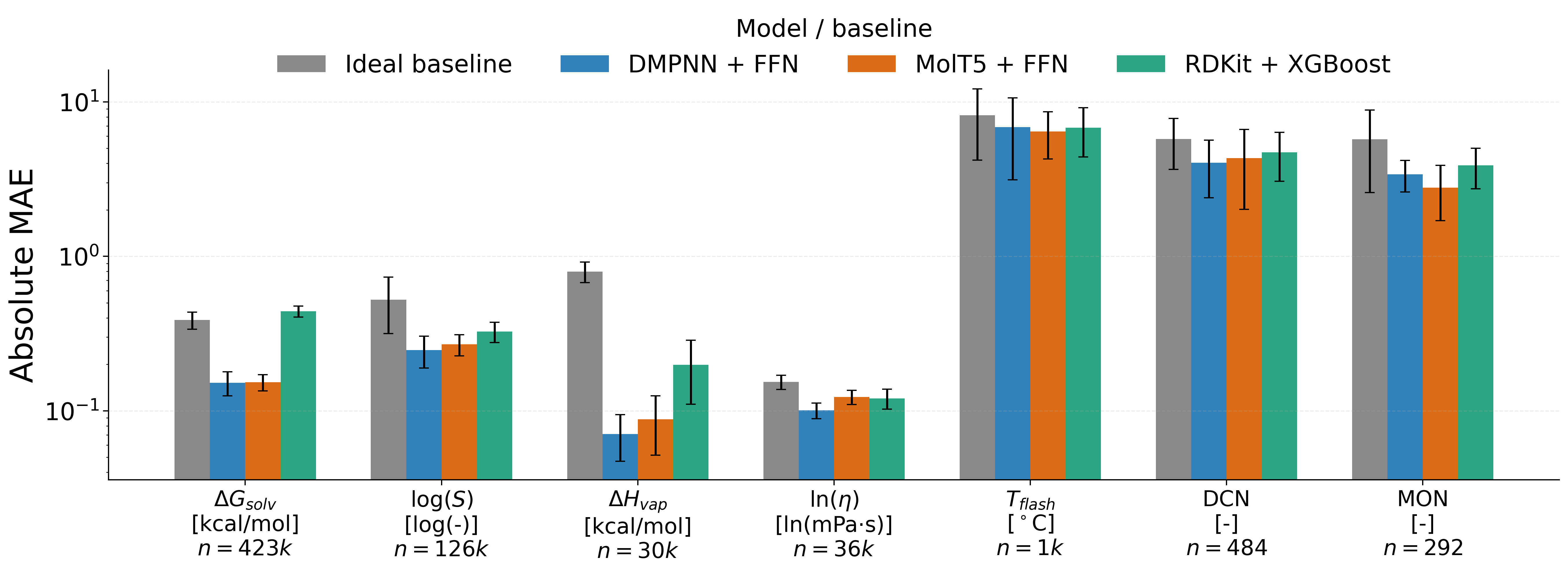}
    \caption{Absolute MAE vs.\ the ideal-mixture reference baseline under the mixture split, for all datasets including $\Delta G_{solv}$ and flashpoint.}
    \label{fig:matrix1-absolute-mae-vs-ideal-baseline-all}
\end{figure}

\begin{figure}[h]
    \centering
    \includegraphics[width=\linewidth]{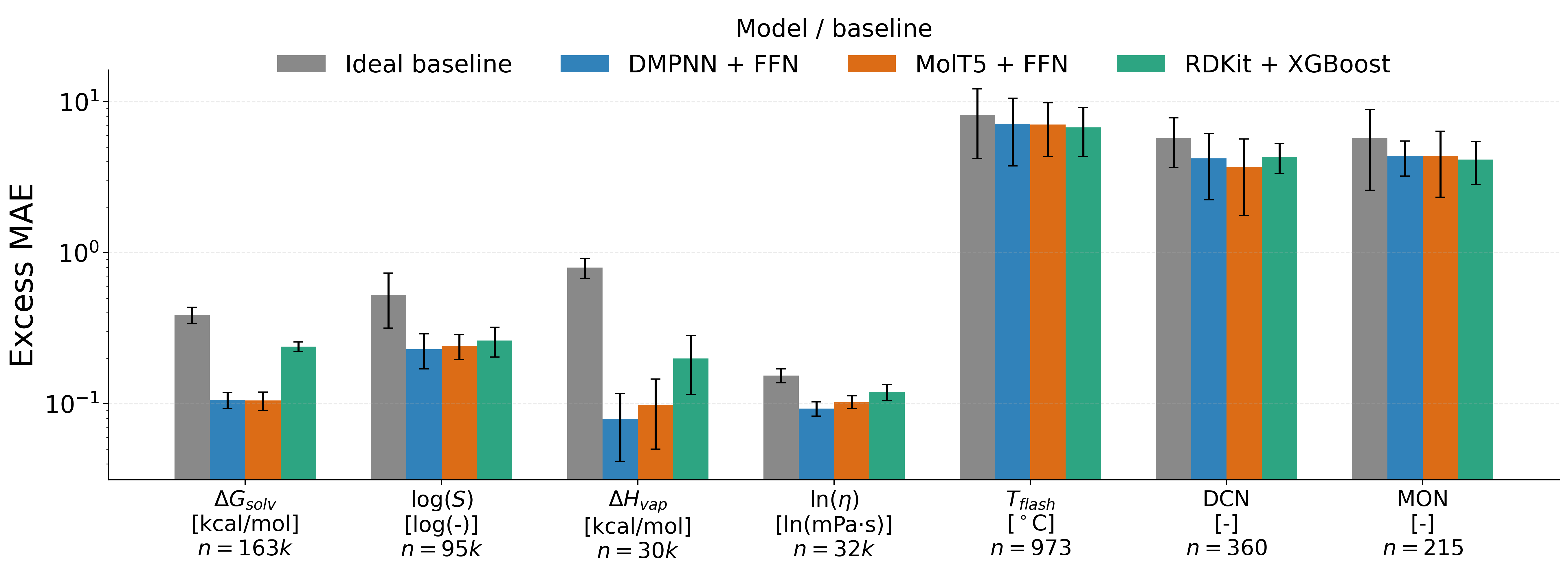}
    \caption{Excess MAE vs.\ the ideal-mixture reference baseline under the mixture split, for all datasets including $\Delta G_{solv}$ and flashpoint. The ideal-baseline bar corresponds to predicting zero excess (a flat mixture), quantifying the non-ideal signal present in each dataset.}
    \label{fig:matrix1-excess-mae-vs-ideal-baseline-all}
\end{figure}

Figures~\ref{fig:matrix1-absolute-mae-vs-ideal-baseline-all} and~\ref{fig:matrix1-excess-mae-vs-ideal-baseline-all} extend the main-text featurization comparison to all datasets, including $\Delta G_{solv}$ and flashpoint, which were omitted from the compact main-text panel for readability. Absolute and excess MAE are shown separately, allowing the contribution of pure-component versus non-ideal behavior to be assessed across the full benchmark.

\subsubsection{Learning curves}
\label{app:learning-curves}
\begin{figure*}[h]
    \centering
    \includegraphics[width=0.44\linewidth]{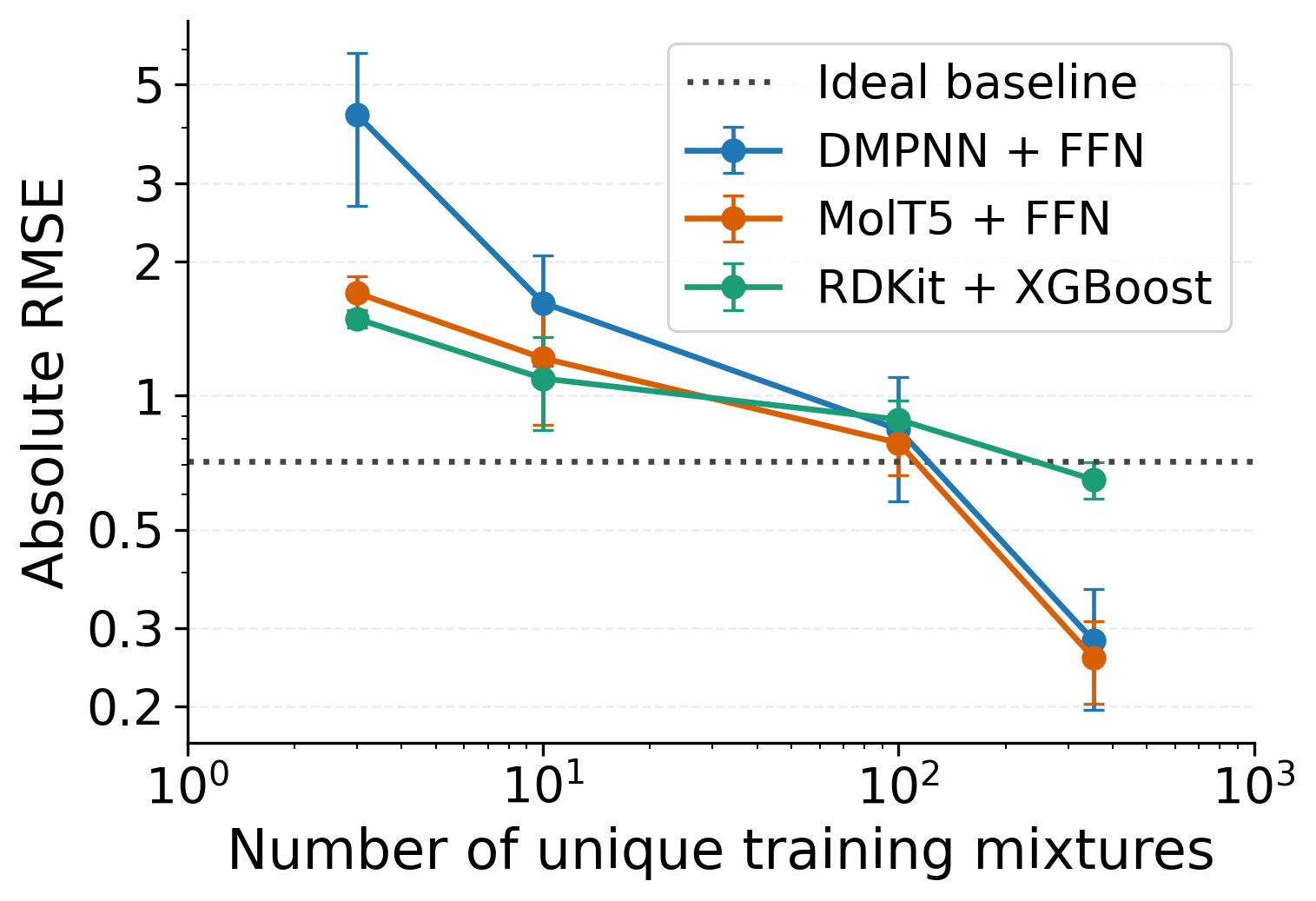}
    \includegraphics[width=0.44\linewidth]{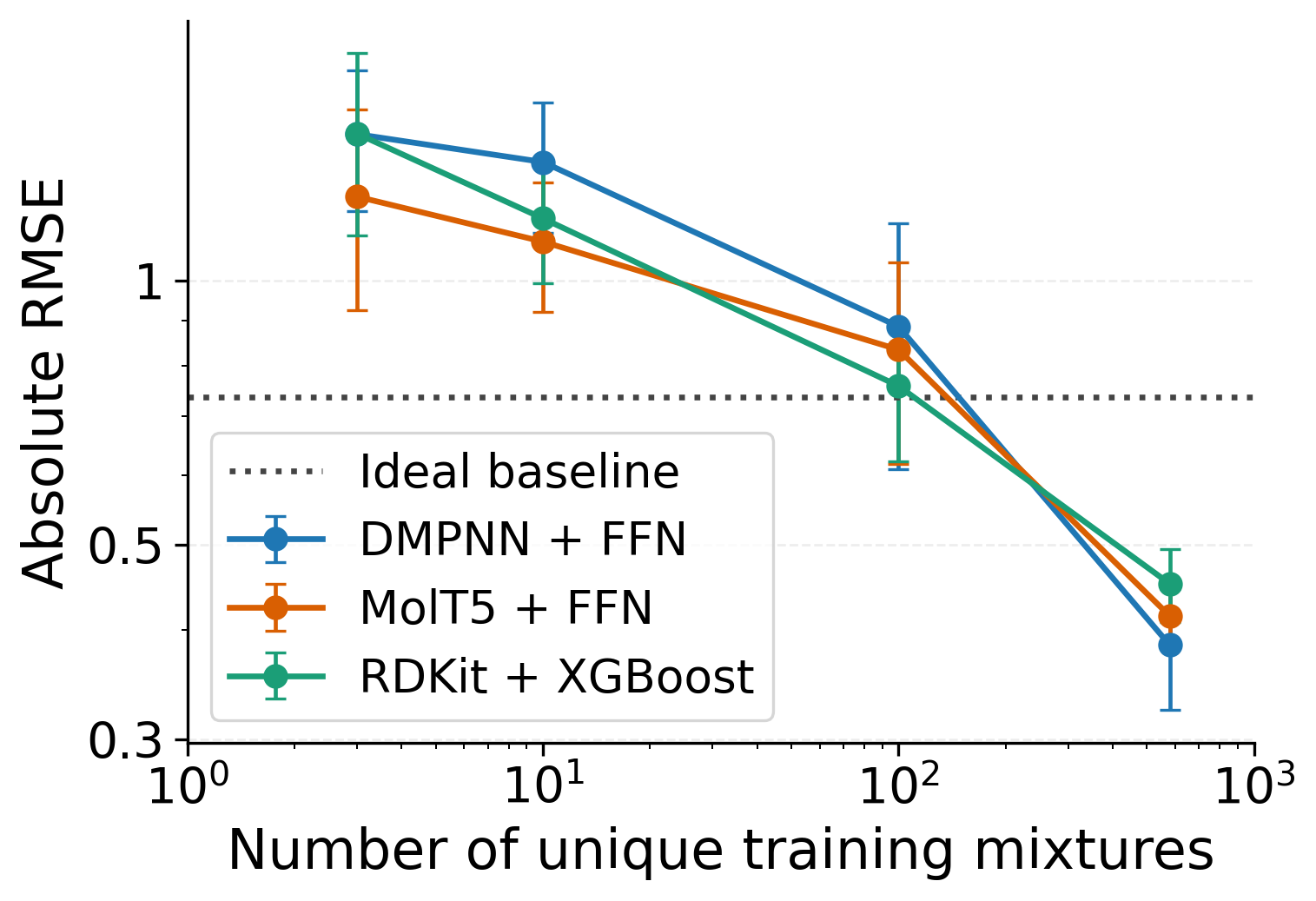}

    \includegraphics[width=0.44\linewidth]{figures/default_hparam_matrix5_viscosity_mix_exp_learning_curve_absolute_rmse.png}
    \includegraphics[width=0.44\linewidth]{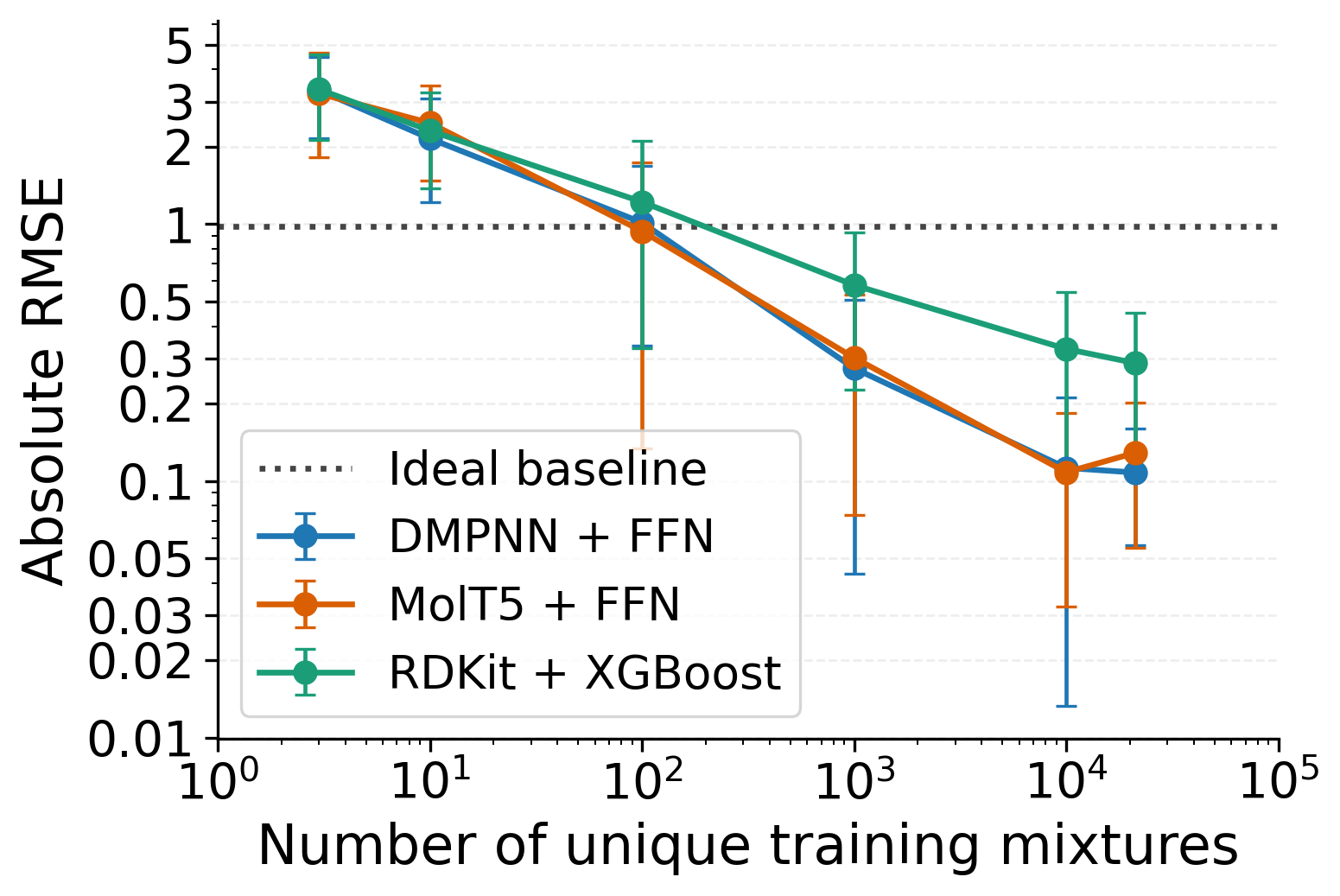}
    \caption{Dataset-wise absolute-RMSE learning curves under the mixture split for $\Delta G_{solv}$, solubility, viscosity, and $\Delta H_{vap}$, using fixed validation and test sets with a progressively downsampled number of unique training mixtures.}
    \label{fig:matrix5-learning-curves-absolute-appendix}
\end{figure*}

\begin{figure*}[h]
    \centering
    \includegraphics[width=0.44\linewidth]{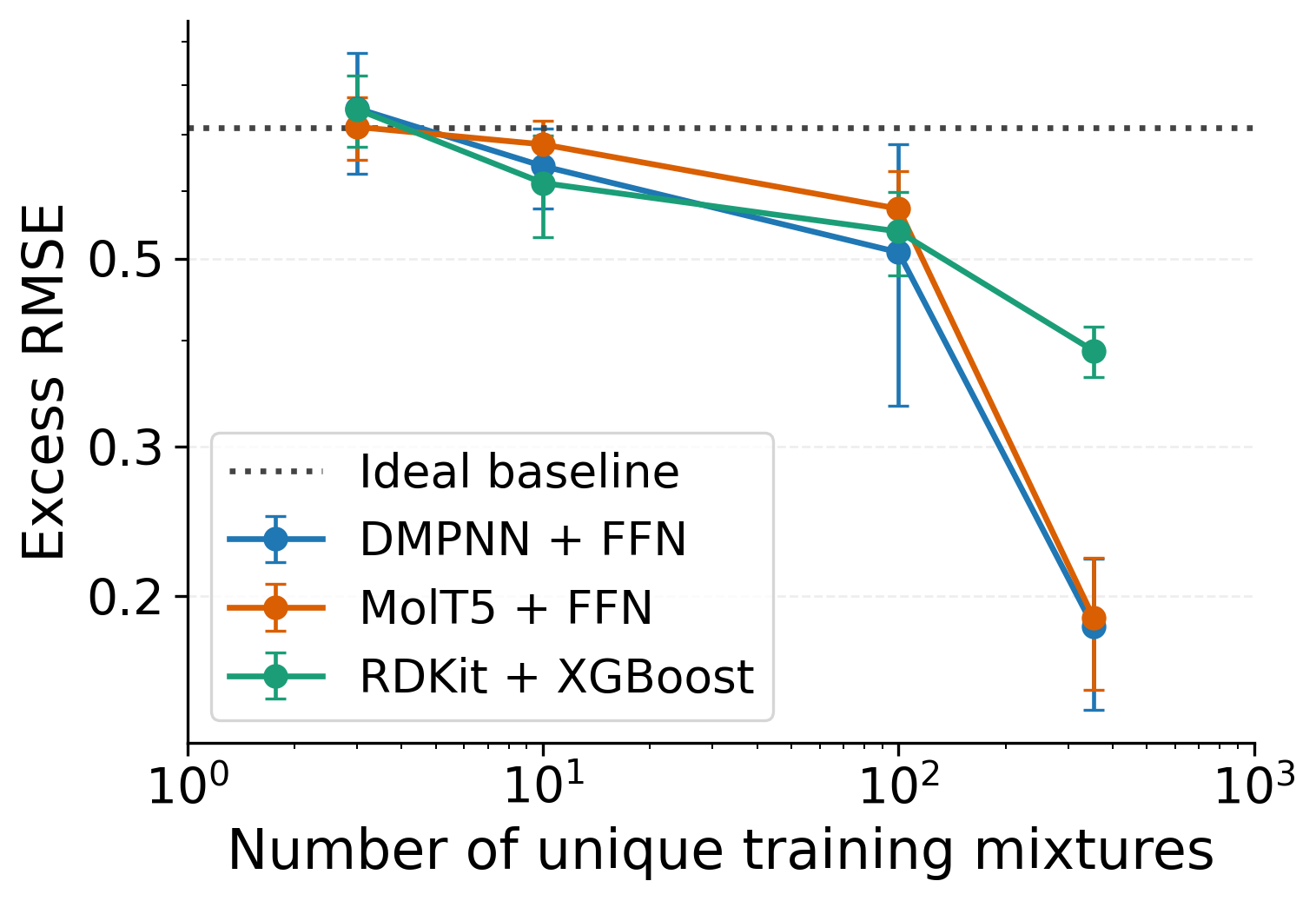}
    \includegraphics[width=0.44\linewidth]{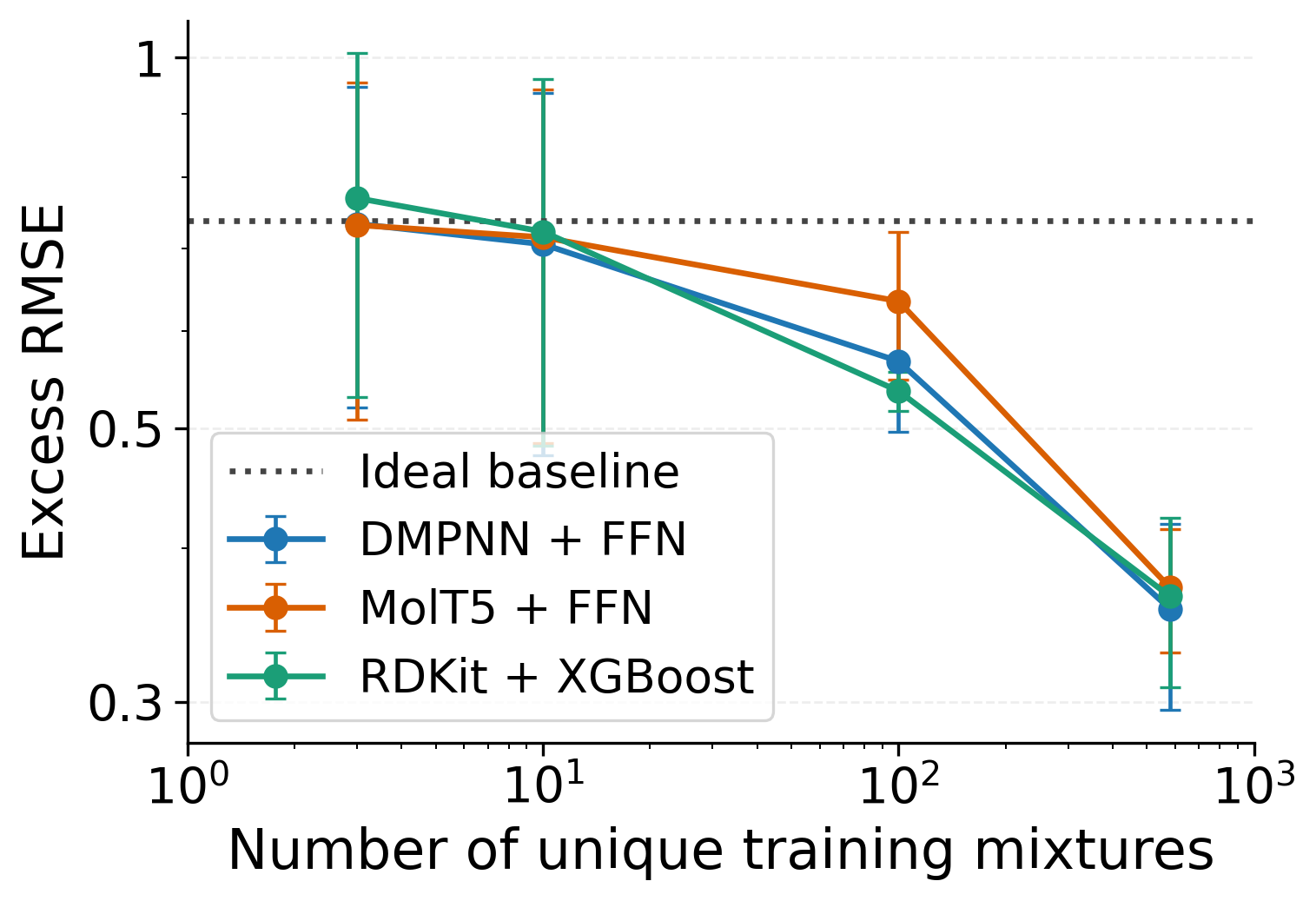}

    \includegraphics[width=0.44\linewidth]{figures/default_hparam_matrix5_viscosity_mix_exp_learning_curve_excess_rmse.png}
    \includegraphics[width=0.44\linewidth]{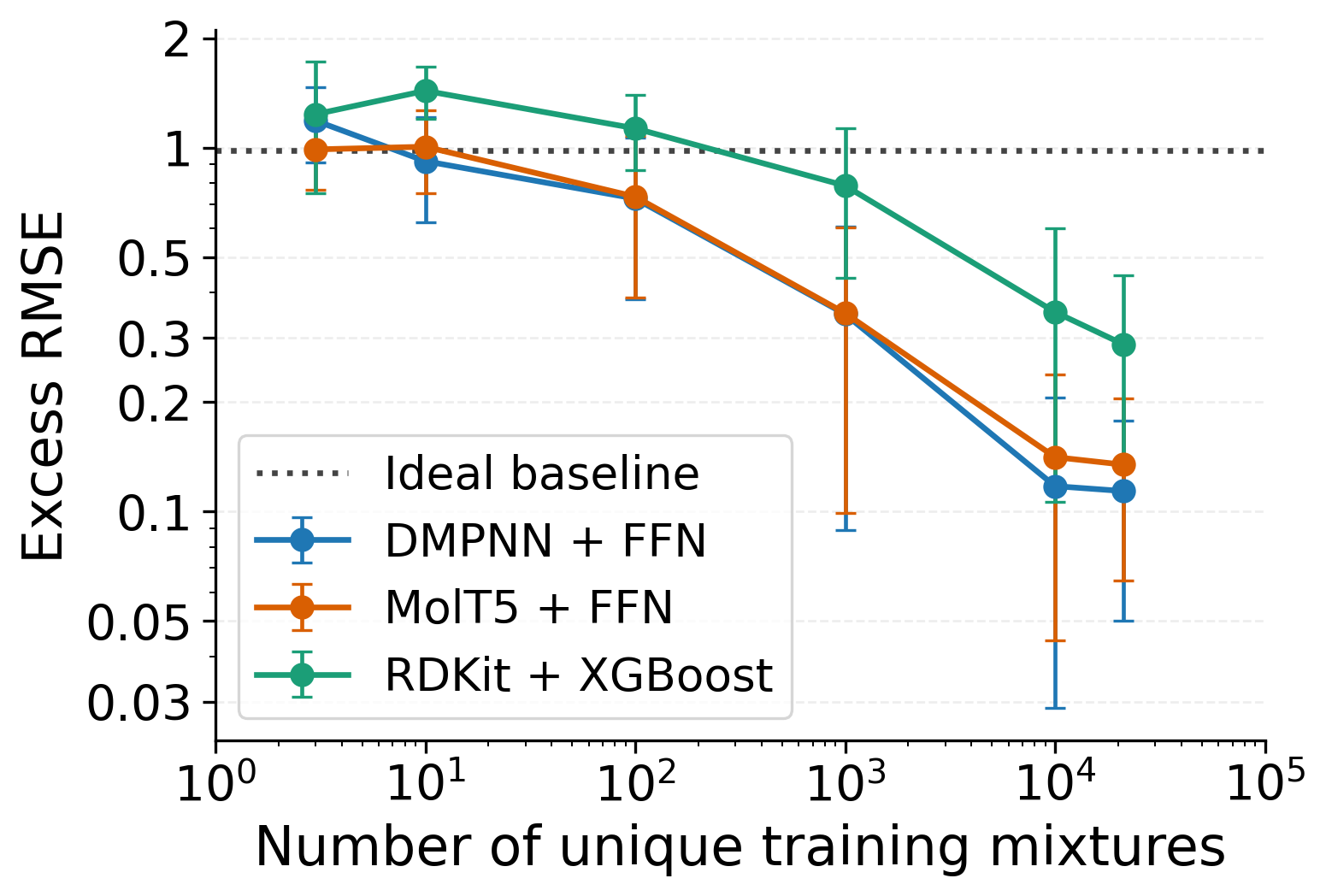}
    \caption{Dataset-wise excess-RMSE learning curves under the mixture split for $\Delta G_{solv}$, solubility, viscosity, and $\Delta H_{vap}$, using fixed validation and test sets with a progressively downsampled number of unique training mixtures.}
    \label{fig:matrix5-learning-curves-excess-appendix}
\end{figure*}

Figures~\ref{fig:matrix5-learning-curves-absolute-appendix} and~\ref{fig:matrix5-learning-curves-excess-appendix} show dataset-wise learning curves under the mixture split, reporting both absolute and excess RMSE as a function of the number of unique training mixtures. These complement the viscosity curves in the main text and illustrate how the data requirement for outperforming the ideal-mixture baseline varies across datasets and metrics.

\subsubsection{Temperature-context diagnostics}
\label{app:temperature-context-diagnostics}
\begin{figure}[h]
    \centering
    \setlength{\tabcolsep}{4pt}
    \begin{tabular}{cc}
        \includegraphics[width=0.48\linewidth]{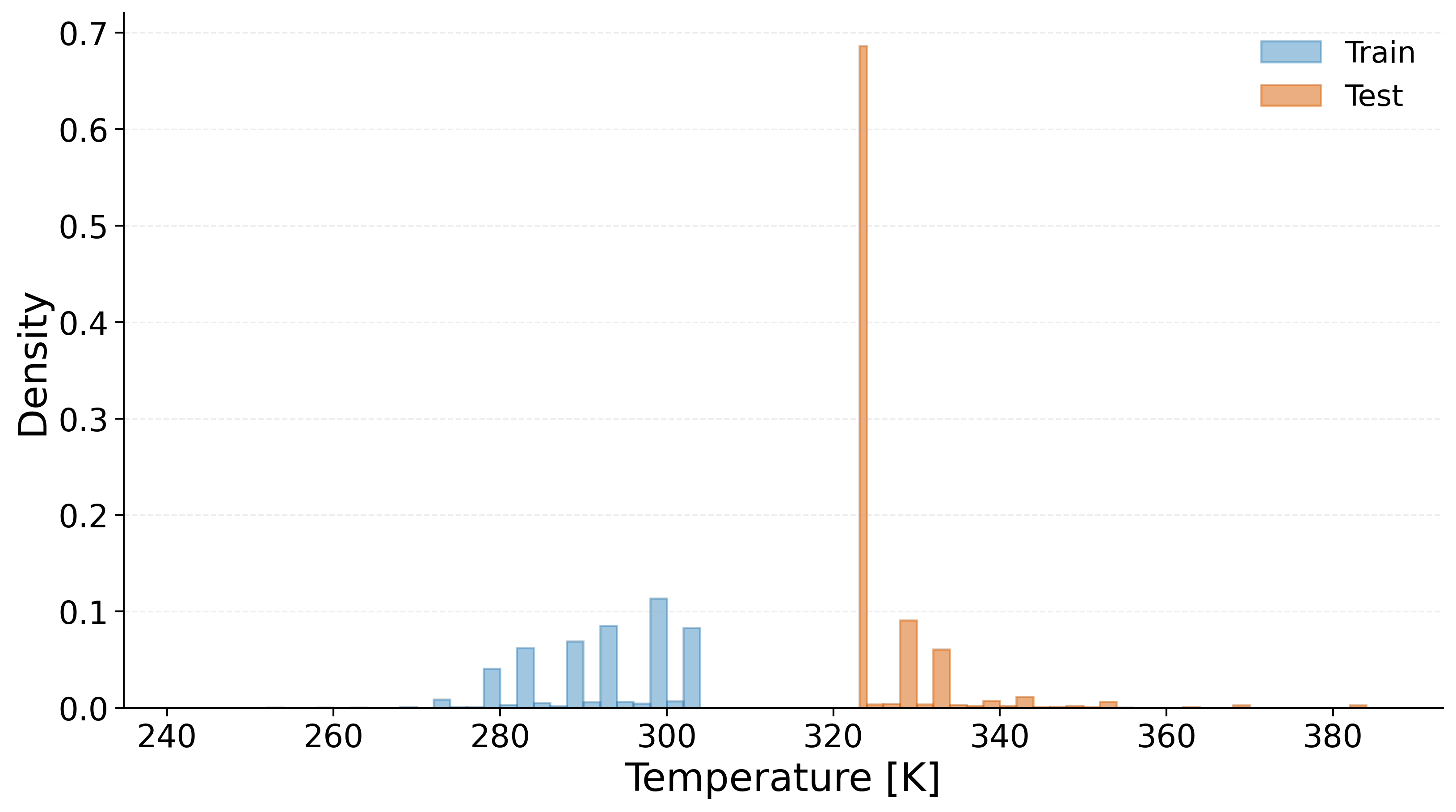} &
        \includegraphics[width=0.48\linewidth]{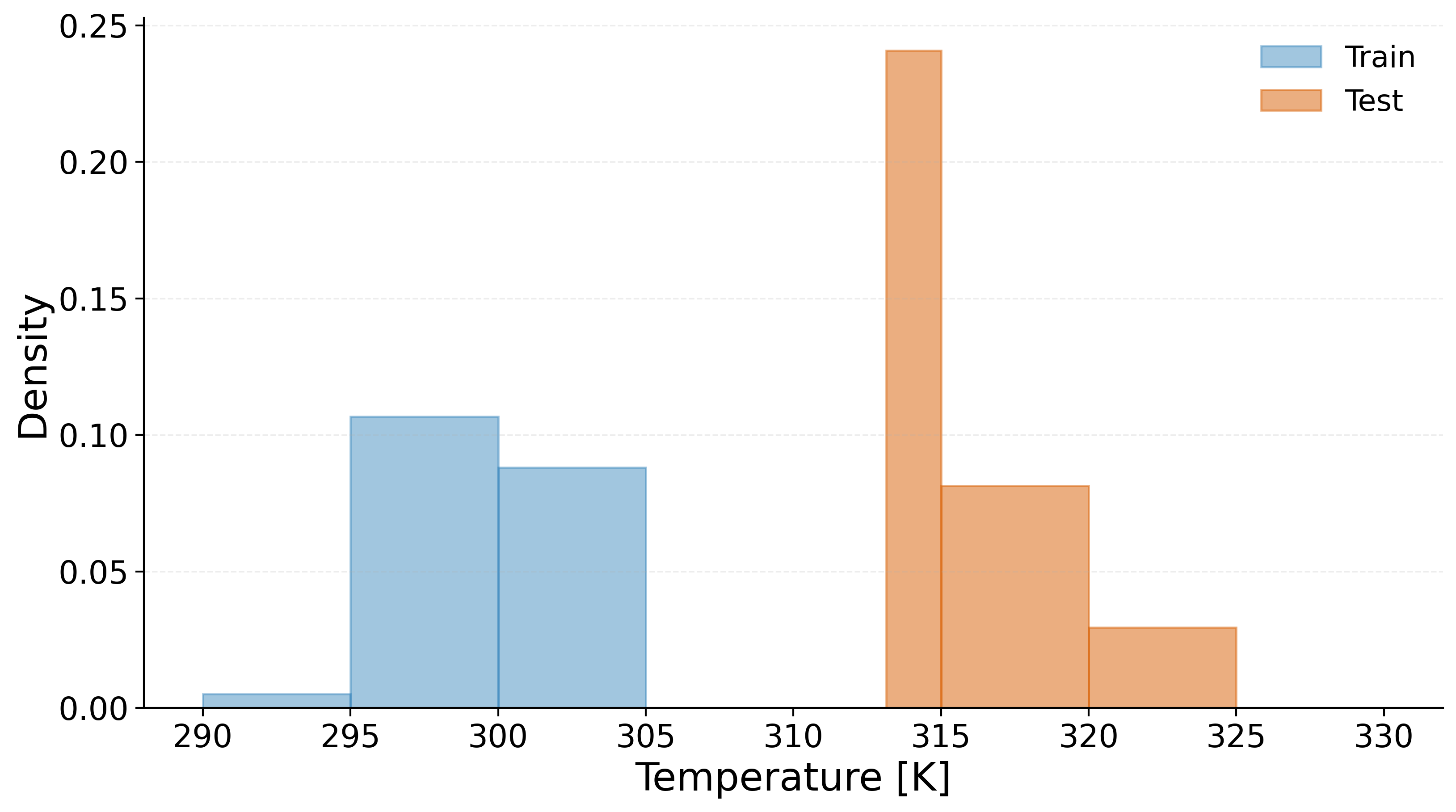}
    \end{tabular}
    \caption{Fold-00 train/test temperature distributions for the solubility and viscosity mixture-temperature splits.}
    \label{fig:temperature-context-fold0-distributions}
\end{figure}

Figure~\ref{fig:temperature-context-fold0-distributions} shows the fold-0 train/test temperature distributions for the solubility and viscosity mixture-temperature splits. These distributions clarify the degree of temperature extrapolation imposed by the split and provide context for interpreting the temperature-context ablation results in the main text.

\subsubsection{Aggregation: MolT5\,+\,FFN}
\label{app:aggregation-molt5}
Table~\ref{tab:pooling-rmse-2split-compact-molt5} reports excess RMSE for MolT5\,+\,FFN across aggregation schemes and splits, complementing the DMPNN\,+\,FFN results in the main text. This allows the reader to assess whether the aggregation trends observed for the graph-based model also hold for the language-model representation.

\begin{table*}[h]
\centering
\scriptsize
\caption{MolT5 + FFN aggregation comparison. Values are excess RMSE $\pm$ std. Bold marks the best per column; underlining marks values within two standard deviations of the best mean.}
\label{tab:pooling-rmse-2split-compact-molt5}
\setlength{\tabcolsep}{5pt}
\resizebox{\textwidth}{!}{%
\begin{tabular}{llcccccc}
\toprule
Split & Aggregation & $\Delta G_{solv}$ & $\log(S)$ & $\ln(\eta)$ & $T_{flash}$ & DCN & MON \\
\midrule
Mixture & attentive & 0.21 $\pm$ 0.031 & \underline{0.38 $\pm$ 0.047} & \underline{0.21 $\pm$ 0.044} & 22. $\pm$ 6.5 & \underline{9.2 $\pm$ 2.3} & \underline{7.4 $\pm$ 1.2} \\
 & weighted-sum & 0.19 $\pm$ 0.033 & \textbf{0.36 $\pm$ 0.057} & \textbf{0.20 $\pm$ 0.045} & \underline{12. $\pm$ 5.2} & \underline{6.9 $\pm$ 3.0} & \textbf{6.6 $\pm$ 3.0} \\
 & DeepSets & \underline{0.16 $\pm$ 0.026} & \underline{0.40 $\pm$ 0.082} & \underline{0.22 $\pm$ 0.060} & \textbf{11. $\pm$ 4.3} & \underline{6.4 $\pm$ 3.4} & \underline{8.0 $\pm$ 3.6} \\
 & Set2Set & \textbf{0.14 $\pm$ 0.013} & \underline{0.40 $\pm$ 0.062} & \underline{0.21 $\pm$ 0.037} & \underline{14. $\pm$ 5.3} & \textbf{6.2 $\pm$ 2.9} & \underline{7.1 $\pm$ 1.7} \\
\midrule
Pure-to-mixture & attentive & \textbf{0.60 $\pm$ 0.17} & \textbf{0.70 $\pm$ 0.19} & 0.41 $\pm$ 0.038 & \underline{27. $\pm$ 14.} & -- & -- \\
 & weighted-sum & \underline{0.64 $\pm$ 0.21} & \underline{0.87 $\pm$ 0.43} & \textbf{0.24 $\pm$ 0.070} & \textbf{13. $\pm$ 8.5} & -- & -- \\
 & DeepSets & \underline{0.86 $\pm$ 0.14} & \underline{0.87 $\pm$ 0.24} & \underline{0.24 $\pm$ 0.060} & \underline{14. $\pm$ 8.0} & -- & -- \\
 & Set2Set & \underline{0.73 $\pm$ 0.19} & \underline{0.73 $\pm$ 0.23} & 0.40 $\pm$ 0.036 & \underline{23. $\pm$ 11.} & -- & -- \\
\bottomrule
\end{tabular}%
}
\end{table*}

\end{document}